\documentclass{article}

\PassOptionsToPackage{numbers, compress}{natbib}

 \usepackage[main, final]{neurips_2026}

\usepackage[utf8]{inputenc} 
\usepackage[T1]{fontenc}    
\usepackage{hyperref}       
\usepackage{url}            
\usepackage{booktabs}       
\usepackage{amsfonts}       
\usepackage{nicefrac}       
\usepackage{microtype}      
\usepackage{xcolor}         
\usepackage{xspace}
\usepackage{multirow}
\usepackage{colortbl}
\usepackage{graphicx}
\usepackage{booktabs}
\usepackage{siunitx}
\usepackage{duckuments}
\usepackage{pifont} 
\usepackage{makecell}
\usepackage{amsthm}
\usepackage{algorithm}
\usepackage[noend]{algpseudocode}
\usepackage{bm}
\usepackage{listings}
\usepackage{bbm}
\usepackage{xcolor}
\usepackage{titletoc}

\title{Handwritten Text Recognition Lives \\ in the High-Pixel Variance Subspace}

\author{
Carlos Garrido-Munoz \quad
Jorge Calvo-Zaragoza \quad \\
University of Alicante, Spain \\
  \texttt{\{carlos.garrido,jorge.calvo\}@ua.es} \\
}

\begin{document}

\maketitle

\begin{abstract}
In self-supervised pretraining for Handwritten Text Recognition (HTR), pixel reconstruction methods tend to outperform contrastive ones, a pattern that sits awkwardly with recent evidence that pixel reconstruction yields uninformative features for natural-image classification. We argue this discrepancy is not an accident but a consequence of how HTR signal is distributed in pixel space. By probing the input distribution directly, we show that HTR's discriminative content is concentrated in the high-variance pixel subspace and is essentially absent from the low-variance one, the inverse of the structure observed for image classification. Under this view, the right SSL family becomes predictable: objectives that preserve high-variance pixel content should transfer best. We test this prediction across six SSL methods spanning three families (pixel-grounded MIM, JEPA-style, and contrastive image--image and image--text), under matched encoder, data, and evaluation protocols on six handwriting benchmarks across five languages. In the full-label setting, pixel-grounded SSL produces the lowest CER on every benchmark and every probe, exposes per-position character information that other families recover only via the readout, and is the only family that benefits from real-data pretraining. Pixel-grounded representations are also more label-efficient. A geometric property of the encoder, its alignment with the high-variance pixel subspace, predicts CER within every SSL method we test. With a pretrained LLM decoder, a frozen pixel-grounded encoder is already competitive with fully fine-tuned supervised baselines, and full fine-tuning beats them on the mean and ranks first or second on every benchmark. These results challenge the prevailing view that pixel reconstruction wastes capacity on irrelevant detail: whether reconstruction is wasted depends on where the discriminative signal lives in the input.
\end{abstract}    
\section{Introduction}
\label{sec:intro}

Handwritten Text Recognition (HTR) aims to transcribe handwritten content into machine-readable text, supporting applications from the digitization of historical archives to form automation and document retrieval. Despite notable advances in recent years~\citep{garrido2025htrsurvey}, performance still hinges on the availability of large quantities of labeled real handwriting, which are expensive to collect and often unavailable for the long tail of historical, regional, and low-resource scripts \citep{garrido_cvpr_2025,selfsupervised_pearrubia_2024}. The vast variability in human handwriting such as writing style, ligatures, slant, ink, paper, and degradation means that systems trained on one writer or one historical period generalize poorly to others, leaving the field perpetually data-hungry~\citep{garrido2025htrsurvey}. Synthetic data is the natural response to this shortage: rendered from a pool of fonts at essentially zero annotation cost, it can be produced in arbitrary quantities and made multilingual by construction. Yet synthetic handwriting fails to reproduce the subtle nuances of real handwriting, and models trained only on synthetic data exhibit a pronounced synthetic-to-real gap that does not close without seeing real samples. This gap is precisely the regime in which representation quality matters most, since no decoder, language prior, or test-time augmentation can recover information the encoder did not learn to expose. 

Self-supervised pretraining (SSL) \cite{he2022masked, lecun2015deep, lecun2022jepa, balestriero2023cookbookselfsupervisedlearning} is the standard tool for closing this kind of gap, with three families dominating the literature: \textbf{(1) reconstruction-based} methods, also known as Masked Image Modeling (MIM), which reconstruct masked content in pixel space~\citep{he2022masked,xie2022simmim}; \textbf{(2) contrastive} methods, which optimize an instance-discrimination loss in either an image--image~\citep{he2020moco,chen2020simclr,chen2021mocov3} or an image--text~\citep{radford2021clip,zhai2023siglip} setting; and \textbf{(3) embedding-predictive architectures}, of which Joint-Embedding Predictive Architectures (JEPA)~\citep{lecun2022jepa,assran2023ijepa} are the canonical instance, predicting the representations of masked regions in a learned latent space.

In HTR, the existing SSL literature does not provide a clean answer to the question of which family is best, nor a principled understanding of why \cite{selfsupervised_pearrubia_2024}. Comparisons across studies are difficult to interpret because pretraining data, evaluation protocols, and probe choices vary substantially across methods~\citep{selfsupervised_pearrubia_2024, zhang2022chaco, aberdam2021seqclr}.
This question matters because the broader self-supervised literature has converged on the intuition that pixel-space reconstruction does not yield representations useful for perception~\citep{balestriero2024reconstruction}: the underlying assumption is that semantic tasks depend on high-level abstractions, and that recovering low-level pixel detail is incidental to that goal: that capacity spent on such detail is capacity wasted. This view has driven the field toward JEPA-style methods \cite{lecun2022jepa, assran2023ijepa, bardes2024vjepa, assran2025vjepa2}, which by construction predict in a learned latent space and discard pixel-level reconstruction altogether. In this paper, we ask whether the same assumption holds for transcription tasks, where the unit of recognition is a character rather than an object category. We use HTR as a testbed and, to do so rigorously, adapt JEPA methods to handwriting for the first time. Our central argument, however, is sharper: HTR is structurally different from semantic understanding tasks, and the signal that determines the transcription lives, paradoxically, in exactly the low-level detail that JEPA discards. The very content the field's dominant philosophy treats as noise is, on HTR, the signal itself. We show that pixel-grounded methods outperform JEPA-style and contrastive ones across all six benchmarks and every SSL protocol we test, and we explain why through a measurable property of the encoder: alignment with the high-variance pixel subspace, which correlates strongly with downstream CER.

\textbf{Contributions:}\newline
\textbf{(C1)} \textbf{A structural account of why pixel-grounded SSL is the right objective for HTR.} The HTR signal lives in the high-variance pixel subspace, the inverse of natural-image classification, and an encoder's alignment with that subspace predicts its probe CER within every SSL method we test.\newline
\textbf{(C2)} \textbf{The first adaptation of JEPA-style methods to text recognition.} We adapt I-JEPA and V-JEPA-2 to HTR as the first test of the JEPA philosophy outside its validated domain.\newline
\textbf{(C3)} \textbf{The first rigorous, line-level SSL comparison for HTR.} We evaluate six SSL methods spanning three families (Pixel-MIM, JEPA-style, Contrastive) under same encoder, data, and evaluation protocols across six benchmarks in five languages, in both synthetic and real pretraining regimes. Findings from this comparison are summarized in the abstract and detailed in Sec.~\ref{sec:experiments}.
\section{Related Work}
\label{sec:related}

\subsection{Handwritten Text Recognition.}
\label{sec:related-word-htr}
HTR has long been dominated by recurrent architectures: bidirectional LSTMs~\cite{Bi-LSTMGRAVES2005602, Hochreiter-LSTM} trained with the Connectionist Temporal Classification (CTC) objective~\cite{connectionist_graves_2006} held the state of the art on standard benchmarks for over a decade~\cite{boosting_aradillas_2021, icdar2017_snchez_2017, icfhr2014_snchez_2014, international_abed_2010}. Attention-based encoder-decoder models~\cite{Bahdanau:ICLR:2015} later emerged as strong alternatives~\cite{attentionhtr_kass_2022, lexicon_kumari_2022, attentionbased_abdallah_2020, endtoend_coquenet_2022}, and a comparative analysis by~\citet{evaluating_michael_2019} provides a detailed account of these sequence-to-sequence approaches. The introduction of Vision Transformers~\cite{Dosovitskiy2020AnII} reshaped the field by enabling more scalable visual encoders, used either in pure encoder-decoder pipelines~\cite{trocr_li_2023, transformerbased_momeni_2023, ocformer_mostafa_2021, transformer_wick_2021} or paired with CTC decoding~\cite{rescoring_wick_2021, dan_coquenet_2023, light_barrere_2022}. Modern HTR systems built on Transformer encoders depend heavily on large labeled corpora for pretraining~\cite{trocr_li_2023, rethinking_diaz_2021}, which is the bottleneck this paper addresses. Self-supervised learning is the standard tool for relaxing this dependence, and a growing body of work has applied it to text recognition~\cite{reading_yang_2022, selfsupervised_pearrubia_2024, aberdam2021seqclr}. The next subsection reviews this literature in detail.

\subsection{Self-supervised pretraining}
\label{sec:related-ssl}

Three families of visual self-supervised pretraining are now standard. \emph{Reconstruction-based} methods, also known as Masked Image Modeling (MIM), reconstruct masked image content either in pixel space~\citep{he2022masked, xie2022simmim} or in a discrete token space~\citep{bao2022beit}. \emph{Embedding-prediction} methods, of which Joint-Embedding Predictive Architectures (JEPA)~\citep{lecun2022jepa} are the canonical instance, predict the representations of masked regions in a learned latent space~\citep{assran2023ijepa, bardes2024vjepa, assran2025vjepa2}. \emph{Contrastive} methods optimize an instance-discrimination loss in either an image--image~\citep{he2020moco, chen2020simclr, chen2021mocov3} or an image--text~\citep{radford2021clip, zhai2023siglip} setting. The relative merits of these families depend strongly on the downstream task. \citet{balestriero2024reconstruction} have recently argued that pixel-space reconstruction concentrates a model's capacity on a subspace of the data that explains pixel variance but is uninformative for perception, making MIM ill-suited to natural-image classification under linear probing. This argument has driven the field toward JEPA-style methods, which by construction predict in a learned latent space rather than in pixel space. We revisit this argument in the HTR setting and find the opposite structure: the high-variance pixel subspace of handwritten line images is precisely where the discriminative signal lives, so the property that hurts MIM on classification is the property that benefits it on transcription.

\subsection{Self-supervised pretraining for text recognition}
\label{sec:related-ssl-tr}
Prior SSL work in text recognition \citep{selfsupervised_pearrubia_2024} has explored three of the four families introduced above. \emph{Pixel-grounded MIM} has been the dominant route for visual SSL in TR: TextDIAE~\citep{souibgui2023textdiae} pretrains a ViT encoder with masking, blurring, and debinarization pretext tasks; DualMAE~\citep{qiao2023dualmae} decouples visual and semantic feature learning with a dual masked autoencoder; MaskOCR~\citep{lyu2023maskocr} uses vertical strip masking to respect the horizontal flow of text lines; and DiG~\citep{yang2022dig} combines SimMIM-style reconstruction with contrastive learning. \emph{Image--image contrastive} methods adapt MoCo-style~\citep{chen2021mocov3} instance discrimination by reorganizing the contrastive unit: SeqCLR~\citep{aberdam2021seqclr} contrasts at the frame level, PerSec~\citep{liu2022persec} at the stroke--semantic level, ChaCo~\citep{zhang2022chaco} at the character level, CMT-Co~\citep{zhang2022cmtco} via character-movement pretext, and RCLSTR~\citep{zhang2023rclstr} via textual relations; SimAN~\citep{luo2022siman} replaces the contrastive objective with a generative one. \emph{Image--text contrastive} pretraining has not been adapted to TR as an SSL objective: existing work uses CLIP only as a frozen pretrained model, either fine-tuned for STR~\citep{zhao2024clip4str} or fused into an existing recognizer~\citep{aberdam2023clipter}. \emph{JEPA-style} pretraining is fully absent: no published work adapts I-JEPA~\citep{assran2023ijepa}, V-JEPA-2~\citep{assran2025vjepa2}, or any embedding-prediction objective to text recognition, in either the scene or handwritten setting.

Our work supplies a structural account, grounded in the variance geometry of handwritten text-line images, and tests it under matched conditions across six methods~\citep{he2022masked, xie2022simmim, assran2023ijepa, assran2025vjepa2, chen2021mocov3, zhai2023siglip}, including the first study of JEPA-style methods on HTR.
\section{HTR signal lives in the top-variance subspace}
\label{sec:variance}

\citet{balestriero2024reconstruction} argue that for natural-image classification, the discriminative signal lives in the low-variance directions of the input pixel distribution, not the high-variance ones. They support this empirically on TinyImageNet, where classifiers trained on the bottom-variance components of the input outperform those trained on the top-variance ones (Fig.~1 of~\citep{balestriero2024reconstruction}). Pixel-space reconstruction objectives are therefore misallocated for classification: the L2 reconstruction loss is dominated by the high-variance pixel directions that the classifier does not need, so the encoder is forced to spend capacity on representations the downstream task discards. This argument has driven recent work toward JEPA-style methods that predict in latent space rather than pixel space~\citep{assran2023ijepa}, on the assumption that the variance structure of natural-image classification is the variance structure of perception.

HTR has the opposite structure. A handwritten line image is almost entirely ink against background, and the discriminative signal is the configuration of strokes; the directions of greatest pixel variance across a dataset are precisely the strokes the recognizer needs to read. We hypothesize that HTR signal lives in the high-variance pixel subspace and is absent from the low-variance one.

To test this, we adapt the projection-then-probe protocol of \citet{balestriero2024reconstruction} from classification to HTR. For each dataset, we compute pixel-PCA on its training images and form two reconstructions per image: one using the smallest set of high-variance principal components (top-K) capturing a fraction $p$ of the total variance, and one using the smallest set of low-variance components matching the same variance budget (bot-K). We train an identical BiLSTM--CTC probe on each reconstruction; full construction details are in App.~\ref{app:pca-construction}.

\begin{figure}[h]
    \begin{minipage}{0.65\linewidth}
    \centering
    \includegraphics[width=\linewidth]{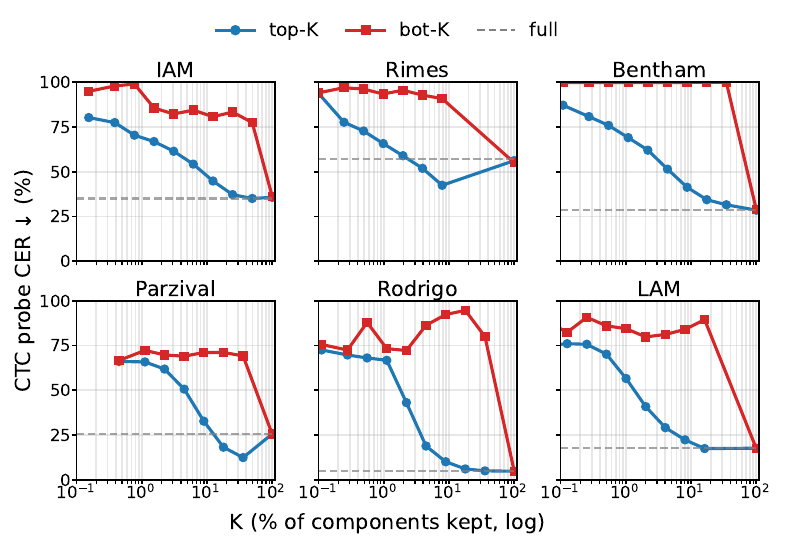}
\end{minipage}\hfill
    \begin{minipage}{0.32\linewidth}
        \caption{\textbf{HTR signal lives in the high-variance pixel subspace.} CTC probe CER as a function of the fraction of pixel principal components retained, on six real handwriting benchmarks. Top-$K$ projections (blue) recover the full-image baseline (dashed) using only the high-variance directions; bot-$K$ projections (red) remain near ceiling. The asymmetry is the inverse of the structure reported by~\citet{balestriero2024reconstruction} for natural-image classification.}
        \label{fig:pca-sweep}
    \end{minipage}
    \vspace{-0.5cm}
\end{figure}

\begin{figure*}[h]
    \centering
    \includegraphics[width=\linewidth]{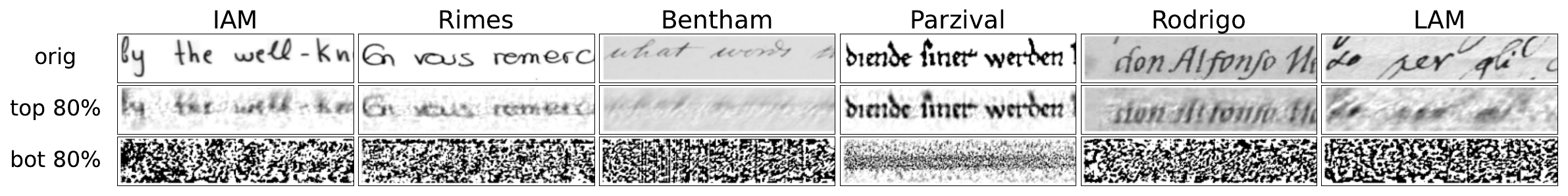}
    \caption{\textbf{Top-$K$ vs.\ bot-$K$ PCA reconstructions of HTR line images.} Per dataset, the top-$K$ row uses the fewest high-variance pixel-PCA components capturing $80\%$ of total variance; the bot-$K$ row uses as many low-variance components as needed to match the same variance budget. Only the top-$K$ reconstruction preserves legible text. This is the inverse of natural-image classification~\citep{balestriero2024reconstruction}.}
    \label{fig:pca-grid}
\end{figure*}

The result reverses the finding of \citet{balestriero2024reconstruction} for natural-image classification: the top-variance probe approaches the full-image baseline once $p$ exceeds a small threshold, while the bot-variance probe remains near ceiling regardless of how much variance is retained (Fig.~\ref{fig:pca-sweep}). Qualitative reconstructions agree: only the top-variance subspace preserves legible text (Fig.~\ref{fig:pca-grid}). The asymmetry is large and consistent across six benchmarks spanning five languages and several historical periods, indicating that handwriting recognition is the inverse of natural-image classification: the discriminative signal concentrates in the high-variance pixel subspace.

The structural account predicts that SSL families whose features preserve high-variance pixel content should transfer better to HTR than families that discard it. The remainder of the paper tests this prediction across six methods drawn from three SSL families.

\section{Methods}
\label{sec:method}

We compare six SSL methods spanning three families, under two pretraining regimes and three evaluation protocols. Loss functions and per-method hyperparameters are reported in the appendix.

\subsection{SSL families}
\label{sec:method-families}

We choose two canonical instances per family rather than test every text-recognition adaptation, since our goal is to characterize the behavior of the underlying SSL \emph{objective} rather than of any particular TR-specific recipe. Each instance below is the foundation that subsequent TR adaptations build on.

\paragraph{Pixel-grounded.} Pixel-grounded methods, also known as Masked Image Modeling (MIM), reconstruct masked image content directly in pixel space. The reconstruction loss is computed pixel-wise, so its gradient is dominated by the directions of largest pixel variance and the encoder is required to retain high-variance content~\citep{balestriero2024reconstruction, balestriero2023cookbookselfsupervisedlearning}. We use MAE~\citep{he2022masked} and SimMIM~\citep{xie2022simmim}: MAE encodes only visible patches and reconstructs masked ones with a lightweight decoder, while SimMIM encodes all patches (masked ones replaced with a learned token) and reconstructs through a linear projection.

\paragraph{JEPA-style.} JEPA-style methods predict the representations of masked regions in a learned latent space rather than in pixel space~\citep{lecun2022jepa, assran2023ijepa}. Because the loss never sees pixels, the encoder can discard high-variance pixel content as long as it is also unpredictable in the latent space. No prior work has adapted I-JEPA, V-JEPA, or any embedding-prediction objective to text recognition. We therefore test the two canonical instances directly: I-JEPA~\citep{assran2023ijepa} and V-JEPA-2~\citep{assran2025vjepa2}. We feed both methods vertical patches of the line image, treating them as the patches (I-JEPA) or frames (V-JEPA-2) of the original input formats, to keep input conditions matched with the other families. Full adaptation details are in Appendix~\ref{app:jepa-adaptation}.

\paragraph{Contrastive (image--image).} Image--image contrastive methods optimize an instance-discrimination loss between two augmented views of the same image~\citep{he2020moco, chen2020simclr}, rewarding invariance to augmentations and suppressing augmentation-induced pixel variance by construction. We use MoCo-v3~\citep{chen2021mocov3} as the canonical modern image--image contrastive recipe. Sequence-aware TR variants such as SeqCLR~\citep{aberdam2021seqclr}, PerSec~\citep{liu2022persec}, ChaCo~\citep{zhang2022chaco}, CMT-Co~\citep{zhang2022cmtco}, and RCLSTR~\citep{zhang2023rclstr} reorganize the contrastive unit (frame, character, subword) but inherit the underlying MoCo-style image--image objective; testing MoCo-v3 directly characterizes that shared foundation.

\paragraph{Contrastive (image--text).} Image--text contrastive methods pull line images with the same transcription toward the same representation, regardless of writer or style~\citep{radford2021clip, zhai2023siglip}. We use SigLIP~\citep{zhai2023siglip}, the only method in our comparison with access to transcription-level supervision during pretraining.

\subsection{Evaluation protocols}
\label{sec:method-protocols}

We evaluate each pretrained encoder under three protocols, reporting Character Error Rate (CER) on the test split of each benchmark, using the best validation checkpoint. The \textbf{visual-only protocols} (Linear--CTC, BiLSTM--CTC) freeze the encoder and read out from its features without a language prior, isolating what the encoder itself exposes. Each probe is trained on the real training split of each benchmark and evaluated on its test split. The \textbf{LLM-based protocol} (LLaVA-style familiy \citep{liu2023llava, lin2024vila, liu2024nvila}) connects the encoder to a pretrained causal language model; we report it under both encoder-frozen training (Stage~1) and full fine-tuning (Stage~2).

\paragraph{Linear+CTC.} A single linear projection from frozen encoder features to character logits, trained with a CTC objective~\citep{connectionist_graves_2006}. This is the strictest probe: with no temporal smoothing or language prior, its CER reflects what the encoder exposes at each spatial position.

\paragraph{BiLSTM+CTC.} A bidirectional LSTM head over frozen features, followed by a linear projection and CTC objective~\citep{connectionist_graves_2006}. The BiLSTM provides sequence-level integration that the linear probe lacks; comparing the two probes (see Sec.~\ref{sec:exp-readout}) tells us how much character information is exposed per-position versus reconstructed by the readout.

\textbf{Pretrained LLM (multi-stage fine-tuning).} A linear projector connects the encoder to a pretrained causal language model. We pretrain the LM from scratch on a multilingual subset of CC100~\citep{conneau-etal-2020-unsupervised-cc100} covering English, French, Italian, Spanish, and German, sized to match the decoder of TrOCR-B~\citep{trocr_li_2023}. Following the standard LP-then-FT recipe of modern decoder-only vision--language models~\citep{liu2023llava, lin2024vila, liu2024nvila}, we train the readout in three stages. In \textbf{Stage~1 (alignment)}, the encoder and the LM are frozen, and only the projector is trained on synthetic image--label pairs to align the visual feature space with the LM's embedding space. In \textbf{Stage~2 (encoder-frozen fine-tuning)}, the projector and the LM are jointly fine-tuned on each real dataset's training split, with the encoder still frozen. In \textbf{Stage~3 (full fine-tuning)}, the encoder is unfrozen and trained jointly with the projector and the LM at a lower learning rate. We report this protocol under two configurations: an \textbf{encoder-frozen} configuration (Stages 1+2) and a \textbf{full fine-tuning} configuration (Stages 1+2+3).

\subsection{Experimental setup}
\label{sec:method-setup}

\paragraph{Encoder.} We use a single ViT-B-scale backbone ($113.55$M parameters) across all six SSL methods. ViT-B is the standard architectural scale in modern SSL benchmarks~\citep{he2022masked,xie2022simmim,assran2023ijepa,zhai2023siglip} and matches the encoder used by TrOCR-B~\citep{trocr_li_2023}, making our results directly comparable to supervised baselines. Following~\citet{lyu2023maskocr}, we use vertical strip patch embedding to match line geometry; this is the only architectural concession to the HTR domain. Implementation details are in Appendix \ref{app:ssl-pretraining}.

\paragraph{Decoder.} For the LLM-based protocol (Sec.~\ref{sec:method-protocols}), we use a single decoder shared across all SSL methods. The decoder is a $200$M-parameter causal LM, sized to match TrOCR-B's, pretrained from scratch on a multilingual subset of CC100~\citep{conneau-etal-2020-unsupervised-cc100} covering the five languages of our benchmarks. Per-stage hyperparameters and training schedules are in Appendix~\ref{app:vlm-pipeline}.

\paragraph{Data.} \textbf{Real handwriting} comes from six benchmarks covering five languages and historical periods from medieval to modern: IAM~\citep{iamdatabase_marti_2002} (English), RIMES~\citep{rimes_2010} (French), Bentham~\citep{bentham_causer2012building} (19th-century English), LAM~\citep{cascianelli2022lam} (Italian), Rodrigo~\citep{serrano-etal-2010-rodrigo} (16th-century Spanish), and Parzival~\citep{parzival_db} (medieval German). We use the canonical line-level train/val/test splits for each benchmark (totals and details in Appendix~\ref{app:datasets}). \textbf{Synthetic line images} are rendered from approximately $3{,}000$ handwritten fonts paired with multilingual text from Project Gutenberg~\citep{gutenberg_dataset}, yielding $2.5$M lines per language balanced across English, French, Italian, Spanish, and German ($12.5$M lines total). Rendering details (font sampling, line-length distribution, augmentation) are in Appendix~\ref{app:datasets}.

\paragraph{Pretraining regimes.} We pretrain each SSL method under two regimes. In the \textbf{synthetic} regime, the encoder is pretrained on the $12.5$M-line synthetic corpus. In the \textbf{real} regime, the encoder is pretrained on the union of the six benchmarks' training splits. Comparing the two regimes tests whether the relative behavior of SSL families is intrinsic to the family or specific to the pretraining distribution.

\section{Results}
\label{sec:experiments}


Pixel-grounded SSL produces the strongest encoder for HTR by every measure we
test, and the structural account of Sec.~\ref{sec:variance} explains why.
We document this through five complementary experiments: a frozen-encoder probe
across families (Sec.~\ref{sec:exp-probe}), an analysis of per-position structure
preservation (Sec.~\ref{sec:exp-readout}), a geometric account that ties encoder
features to CER (Sec.~\ref{sec:exp-mechanism}), a comparison against fully
supervised state-of-the-art HTR baselines (Sec.~\ref{sec:exp-pipeline}), and an
evaluation of label efficiency under limited supervision
(Sec.~\ref{sec:exp-label-efficiency}).

\subsection{Pixel-grounded SSL produces the most readable encoder}
\label{sec:exp-probe}

\begin{figure*}[h]
  \centering
  \includegraphics[width=0.75\textwidth]{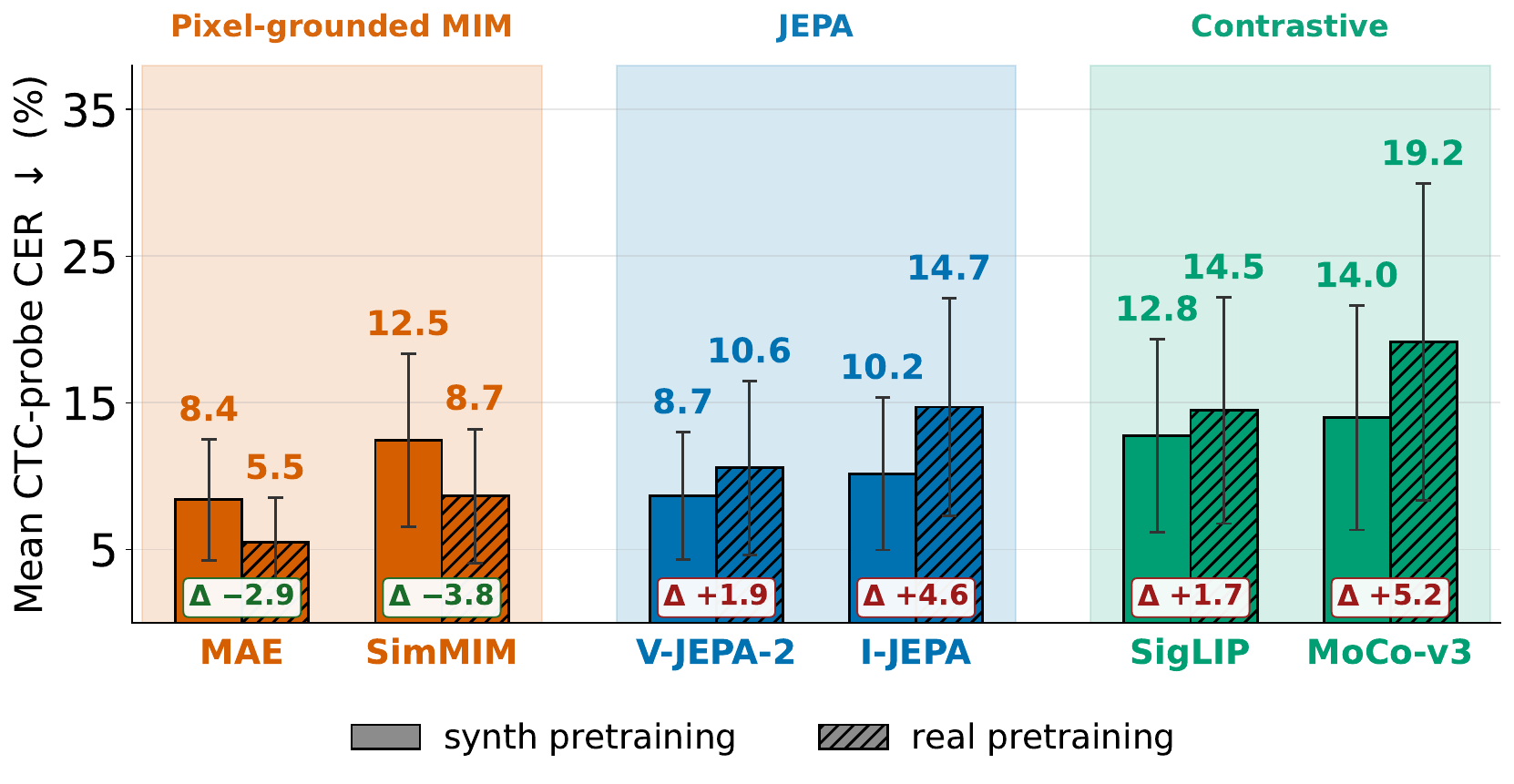}
  \caption{\textbf{Frozen-encoder BiLSTM--CTC probe CER, mean across six benchmarks.} Solid bars: synthetic pretraining; hatched bars: real pretraining. $\Delta$ values report the synth-to-real shift in mean CER. Only pixel-grounded methods improve with real-data SSL; JEPA and contrastive methods degrade. Error bars span the per-benchmark range.}
  \label{fig:probe_cer_summary}
\end{figure*}

\textbf{Pixel-grounded SSL produces the lowest probe CER under both pretraining regimes.} We freeze each pretrained encoder and train a one-layer BiLSTM--CTC probe per benchmark; the probe is the most direct measurement of encoder quality available, since it removes language priors and decoder capacity from the picture. Fig.~\ref{fig:probe_cer_summary} reports the mean test CER across the six real HTR datasets for every method, in both synthetic and real pretraining regimes. Under real pretraining, pixel-grounded SSL is on top: MAE reaches $5.5\%$ mean CER and SimMIM $8.7\%$, followed by V-JEPA-2 ($10.6\%$), then SigLIP ($14.5\%$) and I-JEPA ($14.7\%$), with MoCo-v3 last ($19.2\%$). 

\textbf{Only pixel-grounded SSL benefits from real-data pretraining; every other family degrades.} The $\Delta$ values in Fig.~\ref{fig:probe_cer_summary} report each method's synth-to-real shift in mean CER. MAE and SimMIM improve substantially when SSL is run on real handwriting (MAE: $-2.9$ pp; SimMIM: $-3.8$ pp). The other four methods worsen instead: V-JEPA-2 ($+1.9$ pp), SigLIP ($+1.7$ pp), I-JEPA ($+4.6$ pp), and MoCo-v3 ($+5.2$ pp). The structural account explains the asymmetry: pixel-grounded objectives are aligned with the variance directions that carry the HTR signal, so more real handwriting helps the encoder. The other families' objectives either ignore (JEPA) or actively suppress (contrastive) the high-variance pixel content, and exposing them to more real handwriting at SSL time amplifies the misalignment.

\subsection{Pixel-grounded encoders preserve sequential structure}
\label{sec:exp-readout}

\textbf{Pixel-grounded encoders expose per-position character information; everything else needs the readout to recover it.} A BiLSTM in the readout can recover per-position character information that the encoder did not encode itself, by integrating over the sequence. To separate what the encoder exposes from what the readout reconstructs, we replace the BiLSTM with a single linear layer trained per spatial position (Linear--CTC protocol) and report the gap $\Delta = \text{Linear CER} - \text{BiLSTM CER}$. A small $\Delta$ means the encoder already exposes character information at each position. A large $\Delta$ means the encoder's per-position output is uninformative and the readout must do the work.

\begin{table}[h]
\centering
\caption{Frozen-encoder BiLSTM--CTC test CER ($\%$, $\downarrow$), with Linear--CTC rescue gap $\Delta$ (in pp, in {\color{red}red}). Real-pretrained encoders. \textbf{Bold}\,$=$\,best per column, \underline{underline}\,$=$\,2nd best.}
\label{tab:linear_vs_bilstm_real}
\setlength{\tabcolsep}{3pt}
\renewcommand{\arraystretch}{1.30}
\footnotesize
\begin{tabular}{@{}ll llllll l@{}}
\toprule
Method & Family &
IAM & Rimes & Bentham & LAM & Rodrigo & Parzival & Avg. \\
\midrule
MAE \cite{he2022masked} & Pixel-MIM & \textbf{9.9}\,{\color{red}\scriptsize(+8.8)} & \textbf{6.5}\,{\color{red}\scriptsize(+8.7)} & \textbf{7.5}\,{\color{red}\scriptsize(+6.4)} & \textbf{4.4}\,{\color{red}\scriptsize(+4.1)} & \textbf{2.0}\,{\color{red}\scriptsize(+2.9)} & \textbf{2.5}\,{\color{red}\scriptsize(+3.4)} & \textbf{5.5}\,{\color{red}\scriptsize(+5.7)} \\
SimMIM \cite{xie2022simmim} & Pixel-MIM & \underline{14.5}\,{\color{red}\scriptsize(+8.3)} & \underline{11.1}\,{\color{red}\scriptsize(+9.6)} & \underline{12.0}\,{\color{red}\scriptsize(+7.8)} & \underline{7.0}\,{\color{red}\scriptsize(+12.3)} & \underline{3.0}\,{\color{red}\scriptsize(+2.9)} & 4.3\,{\color{red}\scriptsize(+21.1)} & \underline{8.7}\,{\color{red}\scriptsize(+10.3)} \\
\midrule
I-JEPA \cite{assran2023ijepa} & JEPA & 23.2\,{\color{red}\scriptsize(+34.5)} & 16.9\,{\color{red}\scriptsize(+41.5)} & 22.1\,{\color{red}\scriptsize(+40.4)} & 13.6\,{\color{red}\scriptsize(+43.6)} & 6.0\,{\color{red}\scriptsize(+37.6)} & 6.5\,{\color{red}\scriptsize(+36.3)} & 14.7\,{\color{red}\scriptsize(+39.0)} \\
V-JEPA-2 \cite{assran2025vjepa2} & JEPA & 17.9\,{\color{red}\scriptsize(+26.8)} & 13.2\,{\color{red}\scriptsize(+30.1)} & 15.5\,{\color{red}\scriptsize(+29.4)} & 8.9\,{\color{red}\scriptsize(+24.9)} & 3.8\,{\color{red}\scriptsize(+17.2)} & \underline{4.1}\,{\color{red}\scriptsize(+31.3)} & 10.6\,{\color{red}\scriptsize(+26.6)} \\
\midrule
SigLIP \cite{zhai2023siglip} & Contrastive & 25.1\,{\color{red}\scriptsize(+26.0)} & 16.7\,{\color{red}\scriptsize(+34.7)} & 20.8\,{\color{red}\scriptsize(+33.0)} & 11.3\,{\color{red}\scriptsize(+25.0)} & 5.6\,{\color{red}\scriptsize(+21.5)} & 7.3\,{\color{red}\scriptsize(+36.8)} & 14.5\,{\color{red}\scriptsize(+29.5)} \\
MoCo-v3 \cite{chen2021mocov3} & Contrastive & 32.9\,{\color{red}\scriptsize(+50.7)} & 26.1\,{\color{red}\scriptsize(+63.1)} & 26.0\,{\color{red}\scriptsize(+66.1)} & 15.0\,{\color{red}\scriptsize(+69.9)} & 9.0\,{\color{red}\scriptsize(+58.6)} & 5.8\,{\color{red}\scriptsize(+46.2)} & 19.2\,{\color{red}\scriptsize(+59.1)} \\
\bottomrule
\end{tabular}
\end{table}

The rescue gap separates the families as cleanly as the BiLSTM CER itself. Pixel-grounded encoders need almost no rescue: MAE has a $5.7$ pp average gap, SimMIM $10.3$ pp. JEPA encoders need substantially more (V-JEPA-2: $26.6$ pp; I-JEPA: $39.0$ pp), and contrastive encoders the most (SigLIP: $29.5$ pp; MoCo-v3: $59.1$ pp). MAE's per-position output is the most informative on every benchmark, both in raw CER and in the size of the rescue gap. MoCo-v3's per-position output is the least informative: a linear probe on its frozen features reaches CERs far from the BiLSTM ceiling, and most of the transcription quality on the BiLSTM line is reconstructed by the readout itself. The structural account predicts this directly: an encoder that preserves the high-variance pixel subspace, which carries the stroke content, exposes character information at every spatial position the readout can read.

\subsection{Encoder--subspace alignment explains CER}
\label{sec:exp-mechanism}

We now ask whether the same property explains the encoder-level results: \textit{do encoders whose features carry more of the high-variance pixel content achieve lower CER?} We extend the linear-probing methodology of~\citet{balestriero2024reconstruction} from natural-image classification to HTR encoders. Their analysis projected images onto top-$K$ or bot-$K$ pixel-PCA subspaces and showed that classifiers trained on the bottom subspace outperform those trained on the top. We apply the same projection-then-probe construction at the encoder level: instead of asking how well a classifier reads characters from each subspace, we ask how well the encoder's features themselves can be linearly recovered from each subspace.

\paragraph{Construction of the $R^2$-gap.} For each frozen encoder $\mathcal{E}$ and each dataset $\mathcal{D}$, and for each variance threshold $p \in \{0.10, 0.25, 0.50, 0.75, 0.90\}$, we define matched-variance top and bot pixel-PCA subspaces $V_{\text{top}_p}$ and $V_{\text{bot}_p}$ such that they retain equal total variance ($p$ fraction of the total). Each line image $x$ is projected onto each subspace; we then fit a linear regressor from each projection to the encoder's image-level features and report its coefficient of determination $R^2$. The $R^2$-gap is the differences of the thresholds:
\[
\text{R}^2\text{-gap}(\mathcal{E}, \mathcal{D}) = R^2_{\text{top}_p}(\mathcal{E}, \mathcal{D}) - R^2_{\text{bot}_p}(\mathcal{E}, \mathcal{D}).
\]
A positive gap means the encoder's features are easier to predict from the high-variance pixel directions than from the matched-energy low-variance ones; the encoder geometrically prefers the directions in which the HTR signal lives.

\begin{figure*}[h]
  \centering
  \includegraphics[width=\linewidth]{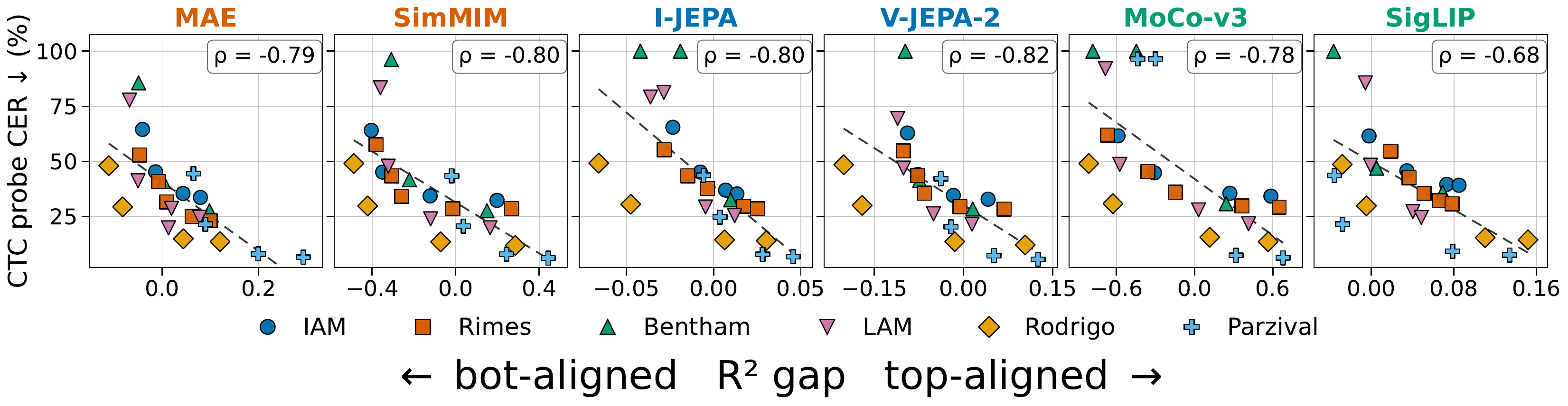}
  \caption{\textbf{Encoder--pixel-variance alignment predicts CER, within every encoder.} For each SSL method (one panel), we plot the per-dataset $R^2$-gap ($R^2_{\mathrm{top}} - R^2_{\mathrm{bot}}$, integrated across thresholds $p$) against the encoder's frozen BiLSTM--CTC probe CER. Marker shape denotes the dataset. Spearman $\rho$ shown per panel; the relationship is strong and consistent across all six methods, including those with poor overall HTR performance.}
  \label{fig:alignment_scatter}
\end{figure*}

\textbf{Better alignment with the high-variance pixel subspace produces lower CER, within every encoder we test.} Fig.~\ref{fig:alignment_scatter} plots the per-dataset $R^2$-gap against the encoder's frozen BiLSTM--CTC probe CER on that dataset, one panel per SSL method. Within every encoder, datasets with higher $R^2$-gap have lower CER, and the relationship is uniformly strong: it holds for the encoders that win at HTR (MAE, SimMIM), for the encoders that lose (MoCo-v3, SigLIP), and for everything in between. Regardless of the SSL objective that produced the features, the alignment of those features with the high-variance pixel subspace tracks downstream HTR performance. This closes the structural account: the encoder property identified in Sec.~\ref{sec:variance} as the operative one for HTR is also the property that empirically predicts an encoder's CER, across families.

\subsection{SSL encoders compete with supervised SOTA HTR}
\label{sec:exp-pipeline}

The probes of Sec.~\ref{sec:exp-probe}--\ref{sec:exp-mechanism} measure properties of the encoder in isolation. They show that pixel-grounded encoders preserve what HTR readouts need, but leave open whether this advantage translates to a complete HTR system competitive with the supervised state of the art. We test this with the LLaVA-style three-stage pipeline of Sec.~\ref{sec:method-protocols} under two configurations: an \textbf{encoder-frozen} configuration (Stage~1 alignment on synthetic data, then Stage~2 fine-tuning of the projector and LM on each real dataset), which isolates the contribution of the SSL representation; and a \textbf{full fine-tuning} configuration that adds Stage~3, unfreezing the encoder and training the entire system. To ensure equal training conditions, all supervised baselines (CRNN, DTrOCR\textsubscript{B}, TrOCR\textsubscript{B}) follow the same data schedule: full supervised training on our $12.5$M-line synthetic corpus, followed by full fine-tuning on each real dataset's training split. Both configurations are reported in Table~\ref{tab:pipeline_sota}.

\textbf{A frozen pixel-grounded encoder is already competitive with supervised SOTA, and full fine-tuning beats it.} Under the encoder-frozen configuration, MAE achieves $5.1\%$ mean CER, lower than CRNN ($5.5\%$), DTrOCR\textsubscript{B} ($5.6\%$), and the no-SSL control ($5.7\%$), and within $0.4$ pp of TrOCR\textsubscript{B} ($4.7\%$), the strongest supervised baseline. The encoder receives no gradient updates at any point in this configuration. Under full fine-tuning, MAE reaches $4.5\%$ mean CER, the lowest result in the table and below TrOCR\textsubscript{B}'s $4.7\%$, and ranks first or second in every column. The gain from the encoder-frozen to the full fine-tuning configuration is small for MAE ($-0.6$ pp on the mean), confirming that its frozen representation was already close to ceiling; SimMIM, JEPA, and contrastive encoders gain more from full fine-tuning ($1.0$ to $1.1$ pp on the mean), consistent with the rescue-gap finding of Sec.~\ref{sec:exp-readout} that their representations require more downstream work to be useful. 

\textbf{The family ordering is preserved across both configurations.} MAE is first under both encoder-frozen and full fine-tuning; SimMIM and V-JEPA-2 sit in the middle; SigLIP is last among SSL methods. The LLM decoder lowers absolute CER uniformly across encoders but does not reorder them, indicating that the language prior cannot recover information the encoder did not expose. The advantage of pixel-grounded SSL is therefore intrinsic to the representation, not an artifact of decoder capacity or training protocol.

\definecolor{baselinegray}{gray}{0.50}
\definecolor{deltagood}{rgb}{0.0,0.55,0.0}
\newcommand{\dgood}[1]{\textsuperscript{\textcolor{deltagood}{(\textminus#1)}}}

\begin{table*}[h]
\centering\small
\setlength{\tabcolsep}{4pt}
\renewcommand{\arraystretch}{0.95}
\caption{\textbf{Test CERs} ($\%$, $\downarrow$). All baselines train on synthetic data, then fine-tune on each real dataset. TrOCR\textsubscript{B} inherits a checkpoint pretrained on $\sim$684M images~\citep{trocr_li_2023}; others start from scratch. SSL encoders are pretrained on real handwriting beforehand. Superscripts: improvement (\textcolor{deltagood}{$-$pp}) over encoder-frozen (stages 1+2). \textbf{Bold}\,$=$\,best, \underline{underline}\,$=$\,2nd.}
\label{tab:pipeline_sota}
\begin{tabular}{l llllll @{\hskip 1.0em} l}
\toprule
Method & IAM & Rimes & Bentham & LAM & Rodrigo & Parzival & Mean \\
\midrule
\multicolumn{8}{l}{{\color{baselinegray}\textsc{Supervised baselines}}} \\
\quad {\color{baselinegray}CRNN \cite{multidimensional_recurrent_puig_2017}}  & {\color{baselinegray}7.4} & {\color{baselinegray}4.9} & {\color{baselinegray}9.6} & {\color{baselinegray}5.6} & {\color{baselinegray}2.8} & {\color{baselinegray}\textbf{2.8}} & {\color{baselinegray}5.5} \\
\quad {\color{baselinegray}DTrOCR\textsubscript{B} \cite{dtrocr_fujitake_2023}} & {\color{baselinegray}7.6} & {\color{baselinegray}4.6} & {\color{baselinegray}8.3} & {\color{baselinegray}4.2} & {\color{baselinegray}2.6} & {\color{baselinegray}6.6} & {\color{baselinegray}5.6} \\
\quad {\color{baselinegray}TrOCR\textsubscript{B} \cite{trocr_li_2023}}  & {\color{baselinegray}\textbf{6.3}} & {\color{baselinegray}\textbf{3.8}} & {\color{baselinegray}\underline{6.9}} & {\color{baselinegray}\textbf{3.5}} & {\color{baselinegray}\underline{2.2}} & {\color{baselinegray}5.5} & {\color{baselinegray}\underline{4.7}} \\
\midrule
\multicolumn{8}{l}{\textsc{NO SSL}} \\
\quad Random init.          & 7.0 & 5.1 & 7.8 & 6.7 & 2.4 & 5.1 & 5.7 \\
\midrule
\multicolumn{8}{l}{\textit{Pixel-MIM}} \\
\quad MAE \cite{he2022masked}    & \underline{6.9}\dgood{0.9} & \underline{4.2}\dgood{0.3} & \textbf{5.6}\dgood{0.4} & \underline{4.0}\dgood{0.2} & \textbf{1.9}\dgood{0.5} & \underline{4.6}\dgood{0.9} & \textbf{4.5}\dgood{0.6} \\
\quad SimMIM \cite{xie2022simmim}  & 12.7\dgood{3.2} & 6.8\dgood{1.0} & 10.8\dgood{1.1} & 6.7\dgood{0.4} & 3.5\dgood{0.4} & 8.2\dgood{0.6} & 8.1\dgood{1.1} \\
\multicolumn{8}{l}{\textit{JEPA}} \\
\quad I-JEPA \cite{assran2023ijepa}   & 18.8\dgood{2.7} & 9.3\dgood{0.3}  & 19.3\dgood{0.7} & 10.4\dgood{0.1} & 5.6\dgood{0.5} & 11.4\dgood{0.3} & 12.5\dgood{0.7} \\
\quad V-JEPA-2 \cite{assran2025vjepa2} & 14.3\dgood{1.7} & 7.8\dgood{0.5}  & 13.7\dgood{1.4} & 8.3\dgood{0.5}  & 4.1\dgood{1.1} & 8.5\dgood{1.2}  & 9.5\dgood{1.0} \\
\multicolumn{8}{l}{\textit{Contrastive}} \\
\quad SigLIP \cite{zhai2023siglip}  & 22.9\dgood{1.2} & 12.1\dgood{0.2} & 20.6\dgood{0.2} & 12.6\dgood{0.1} & 8.1\dgood{0.1} & 13.6\dgood{0.9} & 15.0\dgood{0.4} \\
\quad MoCo-v3 \cite{chen2021mocov3}  & 13.9\dgood{3.8} & 7.5\dgood{0.4}  & 12.0\dgood{0.6} & 5.8\dgood{0.0}  & 3.4\dgood{0.0} & 7.9\dgood{0.6}  & 8.4\dgood{0.9} \\
\bottomrule
\end{tabular}
\end{table*}

\subsection{Pixel-grounded SSL needs fewer labels}
\label{sec:exp-label-efficiency}
We test whether pixel-grounded SSL retains its advantage when transcription labels are scarce. For each encoder pretrained on real handwriting, we freeze its weights and train the same BiLSTM readout with CTC loss on nested subsets containing 1\%, 10\%, 25\%, 50\%, or 100\% of the labeled training data. Figure~\ref{fig:low-label-mean} reports mean test CER across the real HTR benchmarks; per benchmark results appear in Appendix~\ref{app:low-label}.

\begin{figure}[h!]
  \centering
  \vspace{-0.6cm}
  \includegraphics[width=0.5\columnwidth]{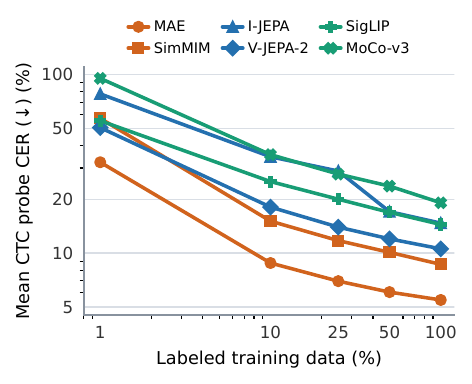}
  \caption{\textbf{Label efficiency of frozen SSL encoders.} Mean test CER
  (\%, $\downarrow$) of probes with a BiLSTM readout and CTC loss trained on
  nested labeled subsets, averaged equally over all real HTR benchmarks.
  Orange denotes pixel reconstruction methods, blue JEPA methods, and green
  contrastive methods; markers distinguish methods within each family. Both
  axes are logarithmic.}
  \label{fig:low-label-mean}
  \vspace{-0.8cm}
\end{figure}

\textbf{Pixel reconstruction remains more label-efficient.}
MAE has the lowest mean CER at every label budget. From 10\% onward, SimMIM
ranks second, so both pixel reconstruction methods outperform the JEPA and
contrastive methods under the same amount of supervision. At 10\% of labels,
MAE reaches 8.8\% mean CER, below the 10.6\% achieved by the strongest JEPA
or contrastive method even with 100\% of labels. The gap therefore reflects
more than an advantage at full supervision: information useful for
transcription can be extracted from pixel reconstruction features with fewer
labeled examples. The consistent advantage across label budgets supports pixel reconstruction as an effective objective for learning HTR representations when transcription labels are limited. 
\section{Conclusion}
\label{sec:conclusion}

Pixel-grounded SSL wins on every measure we test for HTR, and the structural account of Sec.~\ref{sec:variance} explains why: HTR's discriminative signal lives in the high-variance pixel subspace, the inverse of natural-image classification. The pattern is sharpened by an asymmetry in the synth-to-real shift: pixel-grounded methods are the only family that benefits from real-data SSL, while JEPA-style and contrastive methods degrade when given the same real handwriting. The same encoder-level alignment property explains why: the encoder's geometric preference for the high-variance pixel subspace is what makes more real handwriting useful, and it predicts CER within every encoder we test, regardless of family. These results challenge a dominant assumption in self-supervised vision: that pixel reconstruction wastes capacity on detail irrelevant to perception. The assumption holds for tasks whose signal lives in low-variance pixel directions, but not for transcription, where the high-variance stroke detail JEPA-style methods discard is the signal itself. The same argument should apply to other tasks whose discriminative content lives in high-variance pixel directions, such as scene text recognition and music notation recognition; we leave systematic verification to future work. Beyond the structural account, our pipeline produces a state-of-the-art HTR system: under matched training conditions, our MAE-pretrained encoder paired with a pretrained LLM decoder achieves $4.5\%$ mean CER across six benchmarks (vs. TrOCR-B's $4.7\%$) and ranks first or second on every benchmark.

\paragraph{Limitations.} We test a single encoder scale (ViT-B) and a single seed per experiment. Our benchmarks are Latin-script only; whether the structural account holds for non-Latin scripts (Arabic, Chinese, Devanagari) is untested. The account also predicts the same logic for other high-variance-signal transcription tasks (scene text, optical music recognition, mathematical expression recognition), which we do not test directly.
\begin{ack}
This research was supported by the Spanish Ministry of Science and Innovation
through the LEMUR research project (PID2023-148259NB-I00), funded by 
MCIU/AEI/10.13039/501100011033/FEDER, EU, and the European Social Fund Plus
(FSE+). The first author was supported by grant CIACIF/2021/465 from the
Programa I+D+i de la Generalitat Valenciana.
\end{ack}
{
\small
\bibliographystyle{plainnat}
\bibliography{neurips}

@InProceedings{balestriero2024reconstruction,
  title = 	 {How Learning by Reconstruction Produces Uninformative Features For Perception},
  author =       {Balestriero, Randall and Lecun, Yann},
  booktitle = 	 {Proceedings of the 41st International Conference on Machine Learning},
  pages = 	 {2566--2585},
  year = 	 {2024},
  volume = 	 {235},
  series = 	 {Proceedings of Machine Learning Research},
  month = 	 {21--27 Jul},
  publisher =    {PMLR},
}

@String(CVPR  = {IEEE Conf. Comput. Vis. Pattern Recog.})

@String(ICCV  = {Int. Conf. Comput. Vis.})

@String(ICML  = {Int. Conf. Mach. Learn.})

@String(ICLR  = {Int. Conf. Learn. Represent.})

@String(ACCV  = {Asian Conf. Comput. Vis.})

@String(AAAI  = {AAAI})

@String(ICPR  = {Int. Conf. Pattern Recog.})

@String(CVPR  = {CVPR})

@String(ICCV  = {ICCV})

@String(ICML  = {ICML})

@String(ICLR  = {ICLR})

@String(ACCV  = {ACCV})

@String(ICPR  = {ICPR})

@string{anips = {Advances in Neural Information Processing Systems}}

@string{iclr = {International Conference on Learning Representations}}

@string(PMLR = {Proceedings of Machine Learning Research})

@string{aaai = {Proceedings of the AAAI Conference on Artificial Intelligence}}

@string{icml = {International Conference on Machine Learning}}

@string{iccv = {IEEE International Conference on Computer Vision}}

@string{cvpr = {Proceedings of the IEEE conference on Computer Vision and Pattern Recognition}}

@inproceedings{rimes_2010,
  author       = {Christopher Kermorvant and
                  J{\'{e}}r{\^{o}}me Louradour},
  title        = {Handwritten Mail Classification Experiments with the Rimes Database},
  booktitle    = {International Conference on Frontiers in Handwriting Recognition,
                  {ICFHR} 2010, Kolkata, India, 16-18 November 2010},
  pages        = {241--246},
  publisher    = {{IEEE} Computer Society},
  year         = {2010},
  url          = {https://doi.org/10.1109/ICFHR.2010.45},
  doi          = {10.1109/ICFHR.2010.45},
  bibsource    = {dblp computer science bibliography, https://dblp.org}
}

@inproceedings{Bahdanau:ICLR:2015,
  author    = {Dzmitry Bahdanau and
               Kyunghyun Cho and
               Yoshua Bengio},
  editor    = {Yoshua Bengio and
               Yann LeCun},
  title     = {Neural Machine Translation by Jointly Learning to Align and Translate},
  booktitle = {3rd International Conference on Learning Representations, {ICLR} 2015,
               San Diego, CA, USA, May 7-9, 2015, Conference Track Proceedings},
  year      = {2015}
}

@article{Hochreiter-LSTM,
  author       = {Sepp Hochreiter and
                  J{\"{u}}rgen Schmidhuber},
  title        = {Long Short-Term Memory},
  journal      = {Neural Comput.},
  volume       = {9},
  number       = {8},
  pages        = {1735--1780},
  year         = {1997},
  url          = {https://doi.org/10.1162/neco.1997.9.8.1735},
  doi          = {10.1162/NECO.1997.9.8.1735},
  bibsource    = {dblp computer science bibliography, https://dblp.org}
}

@article{Bi-LSTMGRAVES2005602,
title = {Framewise phoneme classification with bidirectional LSTM and other neural network architectures},
journal = {Neural Networks},
volume = {18},
number = {5},
pages = {602-610},
year = {2005},
note = {IJCNN 2005},
issn = {0893-6080},
author = {Alex Graves and Jürgen Schmidhuber},
}

@article{connectionist_graves_2006,
	title        = {Connectionist temporal classification: labelling unsegmented sequence data with recurrent neural networks},
	author       = {Graves, A. and Fernández, Santiago and Gomez, Faustino J. and Schmidhuber, J.},
	year         = 2006,
	journal      = {ICML},
	litmapsid    = 171928617
}

@article{boosting_aradillas_2021,
	title        = {Boosting Offline Handwritten Text Recognition in Historical Documents With Few Labeled Lines},
	author       = {Aradillas, Jose Carlos and Murillo-Fuentes, Juan Jose and Olmos, Pablo M.},
	year         = 2021,
	journal      = {IEEE Access},
}

@article{icdar2017_snchez_2017,
	title        = {ICDAR2017 Competition on Handwritten Text Recognition on the READ Dataset},
	author       = {Sánchez, Joan-Andreu and Romero, Verónica and Toselli, A. and Villegas, M. and Vidal, E.},
	year         = 2017,
	journal      = {2017 14th IAPR International Conference on Document Analysis and Recognition (ICDAR)},
	litmapsid    = 66936741
}

@article{icfhr2014_snchez_2014,
	title        = {ICFHR2014 Competition on Handwritten Text Recognition on Transcriptorium Datasets (HTRtS)},
	author       = {Sánchez, Joan Andreu and Romero, Verónica and Toselli, A. and Vidal, E.},
	year         = 2014,
	journal      = {2014 14th International Conference on Frontiers in Handwriting Recognition},
	litmapsid    = 204945512
}

@article{international_abed_2010,
	title        = {{International Conference on Frontiers in Handwriting Recognition (ICFHR 2010) - Competitions Overview}},
	author       = {Abed, Haikal El and Märgner, Volker and Blumenstein, Michael},
	year         = 2010,
	litmapsid    = 131501335
}

@article{attentionhtr_kass_2022,
	title={AttentionHTR: Handwritten Text Recognition Based on Attention Encoder-Decoder Networks},
  author={D. M. Kass and Ekta Vats},
  journal={ArXiv},
  year={2022},
  volume={abs/2201.09390},
}

@article{lexicon_kumari_2022,
	title        = {Lexicon and attention based handwritten text recognition system},
	author       = {Kumari, Lalita and Singh, Sukhdeep and Rathore, Vaibhav Varish Singh and Sharma, Anuj and Kumari, Lalita and Singh, Sukhdeep and Rathore, Vaibhav Varish Singh and Sharma, Anuj},
	year         = 2022,
}

@article{attentionbased_abdallah_2020,
	title        = {Attention-Based Fully Gated CNN-BGRU for Russian Handwritten Text.},
	author       = {Abdallah, Abdelrahman and Hamada, Mohamed A. and Nurseitov, Daniyar},
	year         = 2020,
	journal      = {Journal of Imaging},

	pubmedid     = 34460538,
	litmapsid    = 261396790
}

@article{endtoend_coquenet_2022,
    author={Coquenet, Denis and Chatelain, Clement and Paquet, Thierry},
journal={ IEEE Transactions on Pattern Analysis \& Machine Intelligence },
title={{ End-to-End Handwritten Paragraph Text Recognition Using a Vertical Attention Network }},
year={2023},
volume={45},
number={01},
ISSN={1939-3539},
pages={508-524},
}

@article{dtrocr_fujitake_2023,
	title        = {DTrOCR: Decoder-only Transformer for Optical Character Recognition},
	author       = {Fujitake, Masato},
	year         = 2023,
	journal      = {arXiv.org},
	litmapsid    = 265628431
}

@article{rethinking_diaz_2021,
	title        = {Rethinking Text Line Recognition Models},
	author       = {Diaz, Daniel Hernandez and Ingle, Reeve and Qin, Siyang and Bissacco, Alessandro and Fujii, Yasuhisa},
	year         = 2021,
	journal      = {arXiv},
	litmapsid    = 261390423
}

@article{Dosovitskiy2020AnII,
	title        = {An Image is Worth 16x16 Words: Transformers for Image Recognition at Scale},
	author       = {Alexey Dosovitskiy and Lucas Beyer and Alexander Kolesnikov and Dirk Weissenborn and Xiaohua Zhai and Thomas Unterthiner and Mostafa Dehghani and Matthias Minderer and Georg Heigold and Sylvain Gelly and Jakob Uszkoreit and Neil Houlsby},
	year         = 2020,
	journal      = {ArXiv},
	volume       = {abs/2010.11929},
}

@article{evaluating_michael_2019,
	title        = {Evaluating Sequence-to-Sequence Models for Handwritten Text Recognition},
	author       = {Michael, Johannes and Labahn, R. and Grüning, Tobias and Zöllner, Jochen},
	year         = 2019,
	journal      = {IEEE International Conference on Document Analysis and Recognition},
	litmapsid    = 195306773
}

@article{rescoring_wick_2021,
	title        = {Rescoring Sequence-to-Sequence Models for Text Line Recognition with CTC-Prefixes},
	author       = {Wick, Christoph and Zöllner, Jochen and Grüning, Tobias},
	year         = 2021,
	journal      = {arXiv: Computer Vision and Pattern Recognition},
	litmapsid    = 138335298
}

@article{dan_coquenet_2023,
	title={DAN: A Segmentation-Free Document Attention Network for Handwritten Document Recognition},
  author={Denis Coquenet and Cl{\'e}ment Chatelain and Thierry Paquet},
  journal={IEEE Transactions on Pattern Analysis and Machine Intelligence},
  year={2022},
  volume={45},
  pages={8227-8243},
}

@InProceedings{light_barrere_2022,
	title        = {A Light Transformer-Based Architecture for Handwritten Text Recognition},
	author       = {Barrere, Killian and Soullard,  Yann  and Lemaitre, Aurélie  and Coüasnon, Bertrand},
	year         = 2022,
	booktitle="Document Analysis Systems",
    publisher="Springer International Publishing",
}

@article{trocr_li_2023,
	title        = {TrOCR: Transformer-Based Optical Character Recognition with Pre-trained Models},
	author       = {Li, Minghao and Lv, Tengchao and Chen, Jingye and Cui, Lei and Lu, Yijuan and Florencio, Dinei and Zhang, Cha and Li, Zhoujun and Wei, Furu},
	year         = 2023,
	journal      = {Proceedings of the ... AAAI Conference on Artificial Intelligence},
	litmapsid    = 260888749
}

@article{transformerbased_momeni_2023,
	title        = {A Transformer-based Approach for Arabic Offline Handwritten Text Recognition},
	author       = {Momeni, Saleh and BabaAli, B.},
	year         = 2023,
	journal      = {arXiv.org},
	litmapsid    = 264978698
}

@article{ocformer_mostafa_2021,
	title        = {OCFormer: A Transformer-Based Model For Arabic Handwritten Text Recognition},
	author       = {Mostafa, Aly and Mohamed, Omar and Ashraf, Ali and Elbehery, Ahmed and Jamal, Salma and Khoriba, Ghada and Ghoneim, A.},
	year         = 2021,
	journal      = {2021 International Mobile, Intelligent, and Ubiquitous Computing Conference (MIUCC)},
	litmapsid    = 213894463
}

@article{transformer_wick_2021,
	title        = {Transformer for Handwritten Text Recognition Using Bidirectional Post-decoding},
	author       = {Wick, C. and Zöllner, Jochen and Grüning, Tobias},
	year         = 2021,
	journal      = {ICDAR},
	litmapsid    = 234797969
}

@article{iamdatabase_marti_2002,
    title = {The IAM-database: an English sentence database for offline handwriting recognition},
  
    author = {Marti, Urs-Viktor and Bunke, Horst},
    journal = {International Journal on Document Analysis and Recognition},
    year = {2002},
    litmapsId = {81093181}
}

@article{lecun2015deep,
  author = {LeCun, Yann and Bengio, Yoshua and Hinton, Geoffrey},
  journal = {nature},
  number = 7553,
  pages = 436,
  publisher = {Nature Publishing Group},
  title = {Deep learning},
  volume = 521,
  year = 2015
}

@inproceedings{multidimensional_recurrent_puig_2017,
  author       = {Joan Puigcerver},
  title        = {Are Multidimensional Recurrent Layers Really Necessary for Handwritten
                  Text Recognition?},
  booktitle    = {14th {IAPR} International Conference on Document Analysis and Recognition,
                  {ICDAR} 2017, Kyoto, Japan, November 9-15, 2017},
  pages        = {67--72},
  publisher    = {{IEEE}},
  year         = {2017},
  doi          = {10.1109/ICDAR.2017.20},
}

@article{reading_yang_2022,
    title = {Reading and Writing: Discriminative and Generative Modeling for Self-Supervised Text Recognition},
  
    author = {Yang, Mingkun and Liao, Minghui and Lu, Pu and Wang, Jing and Zhu, Shenggao and Luo, Hualin and Tian, Qingzhen and Bai, X.},
    journal = {ACM Multimedia},
    year = {2022},
    litmapsId = {255191934}
}

@article{selfsupervised_pearrubia_2024,
  title={Self-Supervised Learning for Text Recognition: A Critical Survey},
  author={Penarrubia, Carlos and Valero-Mas, Jose J and Calvo-Zaragoza, Jorge},
  journal={International Journal of Computer Vision},
  volume={133},
  number={9},
  pages={6221--6250},
  year={2025},
  publisher={Springer}
}

@article{bentham_causer2012building,
  author = {Causer, Tim and Wallace, Valerie},
  title = {Building A Volunteer Community: Results and Findings from Transcribe Bentham},
  journal = {Digital Humanities Quarterly},
  volume = {6},
  number = {2},
  year = {2012},
  publisher = {Providence}
}

@inproceedings{serrano-etal-2010-rodrigo,
    title = "The {RODRIGO} Database",
    author = "Serrano, Nicolas  and
      Castro, Francisco  and
      Juan, Alfons",
    editor = "Calzolari, Nicoletta  and
      Choukri, Khalid  and
      Maegaard, Bente  and
      Mariani, Joseph  and
      Odijk, Jan  and
      Piperidis, Stelios  and
      Rosner, Mike  and
      Tapias, Daniel",
    booktitle = "Proceedings of the Seventh International Conference on Language Resources and Evaluation ({LREC}'10)",
    month = may,
    year = "2010",
    address = "Valletta, Malta",
    publisher = "European Language Resources Association (ELRA)",
  
}

@InProceedings{garrido_cvpr_2025,
    author    = {Garrido-Munoz, Carlos and Calvo-Zaragoza, Jorge},
    title     = {On the Generalization of Handwritten Text Recognition Models},
    booktitle = {Proceedings of the IEEE/CVF Conference on Computer Vision and Pattern Recognition (CVPR)},
    month     = {June},
    year      = {2025},
    pages     = {15275-15286}
}

@inproceedings{liu2023llava,
      title={Visual Instruction Tuning}, 
      author={Liu, Haotian and Li, Chunyuan and Wu, Qingyang and Lee, Yong Jae},
      booktitle=anips,
      year={2023},
}

@inproceedings{cascianelli2022lam,
    title={The LAM Dataset: A Novel Benchmark for Line-Level Handwritten Text Recognition},
    author={Cascianelli, Silvia and Pippi, Vittorio and Martin, Maarand and Cornia, Marcella and Baraldi, Lorenzo and Christopher, Kermorvant and Cucchiara, Rita},
    booktitle=icpr,
    year={2022}
}

@article{garrido2025htrsurvey,
      author={Garrido-Munoz, Carlos and Rios-Vila, Antonio and Calvo-Zaragoza, Jorge},
  journal={IEEE Transactions on Pattern Analysis and Machine Intelligence}, 
  title={Handwritten Text Recognition: A Survey}, 
  year={2025},
  volume={},
  number={},
  pages={1-20},
  doi={10.1109/TPAMI.2025.3646002} 
}

@inproceedings{he2022masked,
  title     = {Masked Autoencoders Are Scalable Vision Learners},
  author    = {He, Kaiming and Chen, Xinlei and Xie, Saining and Li, Yanghao and Doll{\'a}r, Piotr and Girshick, Ross},
  booktitle = {Proceedings of the IEEE/CVF Conference on Computer Vision and Pattern Recognition (CVPR)},
  year      = {2022},
  pages     = {16000--16009}
}

@inproceedings{xie2022simmim,
  title     = {{SimMIM}: A Simple Framework for Masked Image Modeling},
  author    = {Xie, Zhenda and Zhang, Zheng and Cao, Yue and Lin, Yutong and Bao, Jianmin and Yao, Zhuliang and Dai, Qi and Hu, Han},
  booktitle = {Proceedings of the IEEE/CVF Conference on Computer Vision and Pattern Recognition (CVPR)},
  year      = {2022},
  pages     = {9653--9663}
}

@inproceedings{bao2022beit,
  title     = {{BEiT}: {BERT} Pre-Training of Image Transformers},
  author    = {Bao, Hangbo and Dong, Li and Piao, Songhao and Wei, Furu},
  booktitle = {International Conference on Learning Representations (ICLR)},
  year      = {2022}
}

@techreport{lecun2022jepa,
  title       = {A Path Towards Autonomous Machine Intelligence},
  author      = {LeCun, Yann},
  year        = {2022},
  institution = {OpenReview},
  note        = {Version 0.9.2, 2022-06-27},
  url         = {https://openreview.net/forum?id=BZ5a1r-kVsf}
}

@inproceedings{assran2023ijepa,
  title     = {Self-Supervised Learning from Images with a Joint-Embedding Predictive Architecture},
  author    = {Assran, Mahmoud and Duval, Quentin and Misra, Ishan and Bojanowski, Piotr and Vincent, Pascal and Rabbat, Michael and LeCun, Yann and Ballas, Nicolas},
  booktitle = {Proceedings of the IEEE/CVF Conference on Computer Vision and Pattern Recognition (CVPR)},
  year      = {2023},
  pages     = {15619--15629}
}

@article{bardes2024vjepa,
  title   = {Revisiting Feature Prediction for Learning Visual Representations from Video},
  author  = {Bardes, Adrien and Garrido, Quentin and Ponce, Jean and Chen, Xinlei and Rabbat, Michael and LeCun, Yann and Assran, Mahmoud and Ballas, Nicolas},
  journal = {Transactions on Machine Learning Research},
  year    = {2024}
}

@article{assran2025vjepa2,
  title   = {{V-JEPA 2}: Self-Supervised Video Models Enable Understanding, Prediction and Planning},
  author  = {Assran, Mahmoud and Bardes, Adrien and Fan, David and Garrido, Quentin and Howes, Russell and Muckley, Matthew and Rizvi, Aaqib and Roberts, Caner and Sinha, Koustuv and Zholus, Artem and others},
  journal = {arXiv preprint arXiv:2506.09985},
  year    = {2025}
}

@inproceedings{he2020moco,
  title     = {Momentum Contrast for Unsupervised Visual Representation Learning},
  author    = {He, Kaiming and Fan, Haoqi and Wu, Yuxin and Xie, Saining and Girshick, Ross},
  booktitle = {Proceedings of the IEEE/CVF Conference on Computer Vision and Pattern Recognition (CVPR)},
  year      = {2020},
  pages     = {9729--9738}
}

@inproceedings{chen2020simclr,
  title     = {A Simple Framework for Contrastive Learning of Visual Representations},
  author    = {Chen, Ting and Kornblith, Simon and Norouzi, Mohammad and Hinton, Geoffrey},
  booktitle = {International Conference on Machine Learning (ICML)},
  year      = {2020},
  pages     = {1597--1607}
}

@inproceedings{chen2021mocov3,
  title     = {An Empirical Study of Training Self-Supervised Vision Transformers},
  author    = {Chen, Xinlei and Xie, Saining and He, Kaiming},
  booktitle = {Proceedings of the IEEE/CVF International Conference on Computer Vision (ICCV)},
  year      = {2021},
  pages     = {9640--9649}
}

@inproceedings{radford2021clip,
  title     = {Learning Transferable Visual Models from Natural Language Supervision},
  author    = {Radford, Alec and Kim, Jong Wook and Hallacy, Chris and Ramesh, Aditya and Goyal, Gabriel and Agarwal, Sandhini and Sastry, Girish and Askell, Amanda and Mishkin, Pamela and Clark, Jack and Krueger, Gretchen and Sutskever, Ilya},
  booktitle = {International Conference on Machine Learning (ICML)},
  year      = {2021},
  pages     = {8748--8763}
}

@inproceedings{zhai2023siglip,
  title     = {Sigmoid Loss for Language Image Pre-Training},
  author    = {Zhai, Xiaohua and Mustafa, Basil and Kolesnikov, Alexander and Beyer, Lucas},
  booktitle = {Proceedings of the IEEE/CVF International Conference on Computer Vision (ICCV)},
  year      = {2023},
  pages     = {11975--11986}
}

@inproceedings{aberdam2021seqclr,
  title     = {Sequence-to-Sequence Contrastive Learning for Text Recognition},
  author    = {Aberdam, Aviad and Litman, Ron and Tsiper, Shahar and Anschel, Oron and Slossberg, Ron and Mazor, Shai and Manmatha, R. and Perona, Pietro},
  booktitle = {Proceedings of the IEEE/CVF Conference on Computer Vision and Pattern Recognition (CVPR)},
  pages     = {15302--15312},
  year      = {2021}
}

@article{lyu2023maskocr,
  title   = {{MaskOCR}: Text Recognition with Masked Encoder-Decoder Pretraining},
  author  = {Lyu, Pengyuan and Zhang, Chengquan and Liu, Shanshan and Qiao, Meina and Xu, Yangliu and Wu, Liang and Yao, Kun and Han, Junyu and Ding, Errui and Wang, Jingdong},
  journal = {arXiv preprint arXiv:2206.00311},
  year    = {2022}
}

@inproceedings{yang2022dig,
  title     = {Reading and Writing: Discriminative and Generative Modeling for Self-Supervised Text Recognition},
  author    = {Yang, Mingkun and Liao, Minghui and Lu, Pu and Wang, Jing and Zhu, Shenggao and Luo, Hualin and Tian, Qi and Bai, Xiang},
  booktitle = {Proceedings of the 30th ACM International Conference on Multimedia (MM '22)},
  pages     = {4214--4223},
  year      = {2022}
}

@inproceedings{souibgui2023textdiae,
  title     = {{Text-DIAE}: A Self-Supervised Degradation Invariant Autoencoder for Text Recognition and Document Enhancement},
  author    = {Souibgui, Mohamed Ali and Biswas, Sanket and Mafla, Andres and Biten, Ali Furkan and Forn{\'e}s, Alicia and Kessentini, Yousri and Llad{\'o}s, Josep and Gomez, Lluis and Karatzas, Dimosthenis},
  booktitle = {Proceedings of the AAAI Conference on Artificial Intelligence},
  volume    = {37},
  number    = {2},
  pages     = {2330--2338},
  year      = {2023}
}

@inproceedings{liu2022persec,
  title     = {Perceiving Stroke-Semantic Context: Hierarchical Contrastive Learning for Robust Scene Text Recognition},
  author    = {Liu, Hao and Wang, Bin and Bao, Zhimin and Xue, Mobai and Kang, Sheng and Jiang, Deqiang and Liu, Yinsong and Ren, Bo},
  booktitle = {Proceedings of the AAAI Conference on Artificial Intelligence},
  volume    = {36},
  number    = {2},
  pages     = {1702--1710},
  year      = {2022}
}

@inproceedings{zhang2022chaco,
  title     = {{ChaCo}: Character Contrastive Learning for Handwritten Text Recognition},
  author    = {Zhang, Xiaoyi and Wang, Tianwei and Wang, Jiapeng and Jin, Lianwen and Luo, Canjie and Xue, Yang},
  booktitle = {International Conference on Frontiers in Handwriting Recognition (ICFHR)},
  series    = {Lecture Notes in Computer Science},
  volume    = {13639},
  pages     = {345--359},
  publisher = {Springer},
  year      = {2022}
}

@inproceedings{zhang2022cmtco,
  title     = {{CMT-Co}: Contrastive Learning with Character Movement Task for Handwritten Text Recognition},
  author    = {Zhang, Xiaoyi and Wang, Jiapeng and Jin, Lianwen and Ren, Yujin and Xue, Yang},
  booktitle = {Proceedings of the Asian Conference on Computer Vision (ACCV)},
  pages     = {3104--3120},
  year      = {2022}
}

@inproceedings{zhang2023rclstr,
  title     = {{RCLSTR}: Relational Contrastive Learning for Scene Text Recognition},
  author    = {Zhang, Jinglei and Lin, Tiancheng and Xu, Yi and Chen, Kai and Zhang, Rui},
  booktitle = {Proceedings of the 31st ACM International Conference on Multimedia (MM '23)},
  pages     = {5764--5775},
  year      = {2023}
}

@inproceedings{luo2022siman,
  title     = {{SimAN}: Exploring Self-Supervised Representation Learning of Scene Text via Similarity-Aware Normalization},
  author    = {Luo, Canjie and Jin, Lianwen and Chen, Jingdong},
  booktitle = {Proceedings of the IEEE/CVF Conference on Computer Vision and Pattern Recognition (CVPR)},
  pages     = {1039--1048},
  year      = {2022}
}

@misc{balestriero2023cookbookselfsupervisedlearning,
      title={A Cookbook of Self-Supervised Learning}, 
      author={Randall Balestriero and Mark Ibrahim and Vlad Sobal and Ari Morcos and Shashank Shekhar and Tom Goldstein and Florian Bordes and Adrien Bardes and Gregoire Mialon and Yuandong Tian and Avi Schwarzschild and Andrew Gordon Wilson and Jonas Geiping and Quentin Garrido and Pierre Fernandez and Amir Bar and Hamed Pirsiavash and Yann LeCun and Micah Goldblum},
      year={2023},
      eprint={2304.12210},
      archivePrefix={arXiv},
      primaryClass={cs.LG},
      url={https://arxiv.org/abs/2304.12210}, 
}

@inproceedings{conneau-etal-2020-unsupervised-cc100,
    title = "Unsupervised Cross-lingual Representation Learning at Scale",
    author = "Conneau, Alexis  and
      Khandelwal, Kartikay  and
      Goyal, Naman  and
      Chaudhary, Vishrav  and
      Wenzek, Guillaume  and
      Guzm{\'a}n, Francisco  and
      Grave, Edouard  and
      Ott, Myle  and
      Zettlemoyer, Luke  and
      Stoyanov, Veselin",
    editor = "Jurafsky, Dan  and
      Chai, Joyce  and
      Schluter, Natalie  and
      Tetreault, Joel",
    booktitle = "Proceedings of the 58th Annual Meeting of the Association for Computational Linguistics",
    month = jul,
    year = "2020",
    address = "Online",
    publisher = "Association for Computational Linguistics",
    pages = "8440--8451"
}

@inproceedings{lin2024vila,
  title     = {{VILA}: On Pre-training for Visual Language Models},
  author    = {Lin, Ji and Yin, Hongxu and Ping, Wei and Lu, Yao and Molchanov, Pavlo and Tao, Andrew and Mao, Huizi and Kautz, Jan and Shoeybi, Mohammad and Han, Song},
  booktitle = {Proceedings of the IEEE/CVF Conference on Computer Vision and Pattern Recognition (CVPR)},
  pages     = {26679--26689},
  year      = {2024}
}

@inproceedings{liu2024nvila,
  title     = {{NVILA}: Efficient Frontier Visual Language Models},
  author    = {Liu, Zhijian and Zhu, Ligeng and Shi, Baifeng and Zhang, Zhuoyang and Lou, Yuming and Yang, Shang and Xi, Haocheng and Cao, Shiyi and Gu, Yuxian and Li, Dacheng and Li, Xiuyu and Fang, Yunhao and Chen, Yukang and Hsieh, Cheng-Yu and Huang, De-An and Cheng, An-Chieh and Nath, Vishwesh and Hu, Jinyi and Liu, Sifei and Krishna, Ranjay and Xu, Daguang and Wang, Xiaolong and Molchanov, Pavlo and Kautz, Jan and Yin, Hongxu and Han, Song and Lu, Yao},
  booktitle = {Proceedings of the IEEE/CVF Conference on Computer Vision and Pattern Recognition (CVPR)},
  year      = {2025}
}

@INPROCEEDINGS{parzival_db,
  author={Fischer, Andreas and Wuthrich, Markus and Liwicki, Marcus and Frinken, Volkmar and Bunke, Horst and Viehhauser, Gabriel and Stolz, Michael},
  booktitle={2009 15th International Conference on Virtual Systems and Multimedia}, 
  title={Automatic Transcription of Handwritten Medieval Documents}, 
  year={2009},
  volume={},
  number={},
  pages={137-142}}

@article{gutenberg_dataset,
  author       = {Martin Gerlach and
                  Francesc Font{-}Clos},
  title        = {A standardized Project Gutenberg corpus for statistical analysis of
                  natural language and quantitative linguistics},
  journal      = {CoRR},
  volume       = {abs/1812.08092},
  year         = {2018},
}

@article{zhao2024clip4str,
  title   = {{CLIP4STR}: A Simple Baseline for Scene Text Recognition with Pre-trained Vision-Language Model},
  author  = {Zhao, Shuai and Quan, Ruijie and Zhu, Linchao and Yang, Yi},
  journal = {IEEE Transactions on Image Processing},
  volume  = {33},
  pages   = {6893--6904},
  year    = {2024},
  doi     = {10.1109/TIP.2024.3512354}
}

@inproceedings{aberdam2023clipter,
  title     = {{CLIPTER}: Looking at the Bigger Picture in Scene Text Recognition},
  author    = {Aberdam, Aviad and Bensa{\"i}d, David and Golts, Alona and Ganz, Roy and Nuriel, Oren and Tichauer, Royee and Mazor, Shai and Litman, Ron},
  booktitle = {Proceedings of the IEEE/CVF International Conference on Computer Vision (ICCV)},
  pages     = {21706--21717},
  year      = {2023}
}

@inproceedings{qiao2023dualmae,
  title     = {Decoupling Visual-Semantic Features Learning with Dual Masked Autoencoder for Self-Supervised Scene Text Recognition},
  author    = {Qiao, Zhi and Ji, Zhilong and Yuan, Ye and Bai, Jinfeng},
  booktitle = {Document Analysis and Recognition --- ICDAR 2023},
  series    = {Lecture Notes in Computer Science},
  volume    = {14188},
  pages     = {261--279},
  publisher = {Springer},
  year      = {2023},
}
}
\newpage

%
\appendix

\begingroup
\section*{Appendix: Contents}
\setcounter{tocdepth}{1}
\startcontents[appendix]
\printcontents[appendix]{l}{1}{\setcounter{tocdepth}{2}}
\endgroup
\clearpage


This appendix provides the methodological and experimental details
supporting the main paper. We first describe the pixel-PCA construction
used in our variance-based analyses (Sec.~\ref{app:pca-construction}),
followed by the shared SSL pretraining setup and method-specific recipes
(Sec.~\ref{app:ssl-pretraining}). We then introduce the HTR benchmarks
and synthetic data (Sec.~\ref{app:datasets}), the CTC evaluation probes
(Sec.~\ref{app:ctc-probe}), and the evaluation protocol for limited
labeled data (Sec.~\ref{app:low-label}). Finally, we detail the
three-stage vision-language pipeline (Sec.~\ref{app:vlm-pipeline}),
the evaluation protocol and supervised baselines
(Sec.~\ref{app:evaluation}), and the compute and reproducibility
details (Sec.~\ref{app:compute}).

\section{Pixel-PCA construction details}
\label{app:pca-construction}

This appendix describes the image-level pixel-PCA construction
used for the matched-variance top-$K$ and bot-$K$ reconstructions
of Sec.~\ref{sec:variance}, and the separate patch-level
construction used for the $R^2$-gap analysis of
Sec.~\ref{sec:exp-mechanism}.

\paragraph{Setup.}
Let $\mathcal{D}=\{x_i\}_{i=1}^{N}$ denote the training line
images of a dataset, each rescaled to a fixed canvas of
$H=64$ pixels tall by $W=1024$ pixels wide, converted to
grayscale, normalized to $[0,1]$, and flattened into
$x_i\in\mathbb{R}^{d}$ with $d=HW=65{,}536$.
The dataset mean
$\boldsymbol{\mu}=\frac{1}{N}\sum_i x_i$
is computed on the training split only; centering by
$\boldsymbol{\mu}$ before eigendecomposition removes the
mean image shared across observations.

For the $R^2$-gap analysis, each line is instead partitioned
into 256 non-overlapping vertical patches of size
$64\times4$ pixels. Each patch is flattened into a
256-dimensional vector and paired with the corresponding
spatial feature from the frozen encoder, which processes
the complete line.

\paragraph{Eigendecomposition.}
For the image-level analysis, we form the empirical pixel
covariance
\[
\boldsymbol{\Sigma}
=
\frac{1}{N-1}
\sum_{i=1}^{N}
(x_i-\boldsymbol{\mu})(x_i-\boldsymbol{\mu})^\top.
\]
We use the dual formulation based on the $N\times N$
Gram matrix
$G_{ij}=(x_i-\boldsymbol{\mu})^\top(x_j-\boldsymbol{\mu})$.
Its positive-eigenvalue eigenvectors are mapped back to
pixel space to recover the empirical principal directions.
The centered data have rank at most $\min(N-1,d)$.
Writing $r$ for the number of computed components, we obtain
directions $V_r=[v_1,\ldots,v_r]$ and eigenvalues
$\lambda_1\ge\cdots\ge\lambda_r\ge0$.

For the patch-level analysis, we fit a separate PCA basis
by singular value decomposition of up to 500{,}000
training patch vectors after subtracting their pooled mean.
Both bases are fitted exclusively on training data.

\paragraph{Matched-variance subspace selection.}
For the image-reconstruction experiment, let
$\Lambda_{\mathrm{tot}}=\sum_{j=1}^{r}\lambda_j$
denote the total variance of the computed spectrum.
For a target variance fraction $p\in(0,1)$, we select the
smallest top and bot subspaces such that each retains
at least a fraction $p$ of $\Lambda_{\mathrm{tot}}$:
\[
k(p)
=
\min\Bigl\{
K:
\sum_{j=1}^{K}\lambda_j
\ge p\Lambda_{\mathrm{tot}}
\Bigr\},
\qquad
k'(p)
=
\min\Bigl\{
K:
\sum_{j=r-K+1}^{r}\lambda_j
\ge p\Lambda_{\mathrm{tot}}
\Bigr\}.
\]
The corresponding subspaces are
$V_{\mathrm{top}}(p)
=\mathrm{span}\{v_1,\ldots,v_{k(p)}\}$
and
$V_{\mathrm{bot}}(p)
=\mathrm{span}\{v_{r-k'(p)+1},\ldots,v_r\}$.
Their retained variance fractions are approximately
matched, subject to the discrete spectrum, while their
dimensions generally differ.

For the patch-level $R^2$-gap figure, the plotting
implementation uses complementary-variance pairing:
each available top-subspace result retaining variance
fraction $v_{\mathrm{top}}$ is paired with the available
bottom-subspace result whose retained variance fraction
is closest to $1-v_{\mathrm{top}}$.

\paragraph{Reconstruction.}
Each image is reconstructed by projecting the centered
image onto the chosen subspace and re-adding the dataset
mean:
\[
x_p^{\mathrm{top}}
=
\boldsymbol{\mu}
+
\Pi_{V_{\mathrm{top}}(p)}
(x-\boldsymbol{\mu}),
\]
and analogously for $x_p^{\mathrm{bot}}$.
For an orthonormal basis matrix $B$ of a subspace $V$,
$\Pi_V=BB^\top$ is its orthogonal projector.
The full reconstruction uses all computed components.
Reconstructions are clipped to $[0,1]$ before being fed
to the downstream probe.

\paragraph{Thresholds.}
The variance-sweep figure (Fig.~\ref{fig:pca-sweep})
reports probe CER at
$p\in\{0.05,0.10,0.20,0.30,0.50,0.70,0.90\}$;
the qualitative reconstructions
(Fig.~\ref{fig:pca-grid}) use $p=0.80$.
The $R^2$-gap analysis of
Sec.~\ref{sec:exp-mechanism} uses
$p\in\{0.10,0.25,0.50,0.75,0.90\}$,
with each threshold evaluated separately,
without averaging across thresholds.

\paragraph{Probe training.}
For every image-reconstruction setting, we train a
freshly initialised one-layer BiLSTM--CTC head using the
real image-label pairs, following the training protocol
of Sec.~\ref{app:ctc-probe}.
Each reconstructed image is presented as a sequence of
256 flattened $64\times4$-pixel patches.
The full reconstruction provides the reference condition.

For the patch-level encoder analysis, we fit an affine
ridge regressor from the selected pixel-PCA coordinates
to the corresponding frozen encoder features, using up
to 200{,}000 training patches.
Regression quality is measured by $R^2$ on up to
200{,}000 validation patches, pooling squared errors
across patch observations and feature dimensions.
The gap is
$\Delta R^2=R^2_{\mathrm{top}}-R^2_{\mathrm{bot}}$
for the paired subspace results.
We replace the original encoder features with their
ridge predictions and train a fresh BiLSTM--CTC probe
on the resulting sequences.
Each point in the $R^2$-gap figure represents a dataset
and a top-subspace setting: its horizontal coordinate
is the paired validation $\Delta R^2$, and its vertical
coordinate is the test CER of the probe trained on the
corresponding top-subspace ridge-predicted features.
Validation and test data do not enter PCA fitting or
ridge-regression fitting.

\section{SSL encoder pretraining}
\label{app:ssl-pretraining}

We pretrain six self-supervised encoder methods representing three different families using the same backbone and input pipeline, with method-specific adjustments to the optimization setup. We first describe the shared components and then the per-family training recipes.

\paragraph{Image preprocessing.}
All line images are resized while preserving aspect ratio to fit within a
$64 \times 1024$ canvas, padded on the right with white to fill the canvas
(left-aligned), and converted to grayscale. The fraction of lines longer than
$1024$ pixels at $H = 64$ is below $0.5\%$ across all six datasets; these
lines are truncated at the right edge of the canvas.

\paragraph{Training-time augmentation.}
On real handwriting pretraining and on supervised stages we apply a fixed
augmentation pipeline at training time only. The pipeline mixes geometric
perturbations (small random affine transforms covering rotation, translation,
scale, and shear; an occasional perspective warp; a low-probability elastic
deformation, with white fill on introduced regions), pixel-level perturbations
(mild Gaussian blur, brightness and contrast jitter, and low-amplitude
additive Gaussian noise), random ink dilation or erosion with small kernels
applied at low probability, and a low-probability cutout that erases small
rectangular regions and fills them with the page background. Validation and
test sets receive no augmentation.

\paragraph{Backbone.}
Each encoder is a Vision Transformer \cite{Dosovitskiy2020AnII} that consumes full-height vertical
patches as in \cite{lyu2023maskocr} of width $p_w = 4$ pixels (rather than the usual square patches),
so each patch covers
$H \times p_w = 64 \times 4 = 256$ pixels and a line yields
$T = W / p_w = 256$ patches. Position information is provided by 1D rotary
position embeddings along the time axis. The encoder is QK-Norm, LayerScale,
and SwiGLU stabilised, and we use no class token; image-level features (used
by SigLIP and MoCo-v3 for their projection heads) are obtained by mean-pooling
across the $256$ patch tokens. 
The MLP ratio is $4.0$ throughout, and unless otherwise stated the paper
reports the base size.

\paragraph{Vocabulary.}
The character vocabulary consists of $94$ characters (Latin letters,
digits, and common punctuation) shared across all six datasets, plus a
single CTC blank symbol.

\paragraph{Pretraining data.}
We use two regimes. The \emph{real} regime is a balanced union of the six
real HTR datasets (IAM, Rimes, Bentham, LAM, Rodrigo, Parzival) sampled with a
low-temperature schedule so each dataset contributes equally. Supervised baselines do
not have an analogous pretraining stage. The \emph{synth} regime consists of pre-generated
synthetic handwriting line images at the same resolution, rendered from
Gutenberg \cite{gutenberg_dataset} text in five languages
(English, Spanish, French, German, Italian) with the same temperature-balanced
schedule across languages; each language contributes roughly two and half million
lines, for 12.5 million in total. Both regimes use the same image preprocessing and training budget,
with method-specific optimization settings as described below.
Data augmentation is applied to real handwriting, while synthetic
images incorporate variation during generation.

\paragraph{Common optimizer recipe.}
We optimise with AdamW ($\beta_1 = 0.9$, $\beta_2 = 0.95$, weight decay $0.1$),
gradient clipping at norm $5.0$, and a linear warmup of $10{,}000$ steps to a
peak learning rate followed by cosine decay to zero over the remaining steps.
The peak learning rate is $3 \times 10^{-4}$ for masked-image modeling and
JEPA families, and $6 \times 10^{-4}$ for the contrastive SigLIP family. The
total budget is 500 epochs of $2{,}000$ optimiser steps each, totalling
roughly $10^{6}$ updates. Effective batch size is 32 for the masked-image modeling and JEPA
methods, 256 for MoCo-v3, and 1024 for SigLIP, using gradient
accumulation when necessary. These method-specific batch sizes
accommodate the requirements of contrastive learning while
preserving the shared backbone and optimization setup.
All training is in bfloat16 mixed precision.

\paragraph{Checkpoint selection.}
We checkpoint by best validation of an inline encoder-quality CTC probe
trained on IAM features and evaluated every few epochs; the released
checkpoint corresponds to the best inline probe CER seen during training.
This inline probe serves as a stopping/checkpointing signal during
pretraining only; downstream evaluation across all six benchmarks happens
with separate per-dataset probes (Sec.~\ref{app:ctc-probe}) and the VLM
pipeline (Sec.~\ref{app:vlm-pipeline}).

\subsection{MAE.}
\label{app:mae-recipe}
MAE is the canonical pixel-grounded masked image modeling family
\citep{he2022masked}: predict the missing patches of an image in pixel space.
The encoder is the shared vertical-patch ViT described above.
The decoder is a small four-block transformer with hidden dimension 256 and
eight attention heads (4.5M parameters at base), and it predicts the pixel
intensities of the masked patches. We use random patch masking at a 75\% rate, so the encoder processes
only the visible 25\% of patches while the decoder receives the encoder
outputs together with learned mask tokens at the masked positions. The
training objective is mean squared error on the pixel block of each masked
patch only; we do not normalise target pixels, predicting raw intensities
directly.

\subsection{SimMIM.}
\label{app:simmim-recipe}
SimMIM is also a pixel-grounded masked image modeling family
\citep{xie2022simmim}, but with a lightweight single-layer linear projection
in place of MAE's transformer decoder. The encoder is the same vertical-patch
ViT, and in contrast to MAE the masked patches are kept at the encoder input
as learnable mask tokens. The mask predictor is a single linear projection
from encoder hidden states to pixel intensities of each masked patch. We use
random patch masking at a 60\% rate and an L1 reconstruction loss on masked
patches only.

\subsection{I-JEPA (and JEPA adaptation to HTR).}
\label{app:ijepa-recipe}
\label{app:jepa-adaptation}
I-JEPA is a joint-embedding predictive architecture \citep{assran2023ijepa}:
it predicts feature-space representations of masked target patches from
feature-space representations of an unmasked context. The original formulation targets square natural images with a 2D patch grid, samples multiple potentially overlapping target blocks as random rectangles
in 2D, and samples a single context block per image. None of these choices transfers cleanly to long, single-row
line images, where the only meaningful axis is left-to-right and where
2D rectangular masks would span multiple words horizontally while leaving
full character height visible at the top and bottom of the image. We adapt
I-JEPA to HTR along three axes: input geometry, masking strategy, and
loss aggregation.

\textbf{Input geometry.} We treat each line image as a 1D sequence of
$T = 256$ full-height vertical patches (Sec.~\ref{app:ssl-pretraining}),
not as a 2D grid; the encoder, the target encoder, and the predictor all
operate on a single time axis with 1D rotary positional encodings. There
is no 2D positional grid, no class token, and no separate spatial pooling.
Compared to the original 2D I-JEPA, we drop both the global-context
prefix (the original samples a large context block then crops, which is
ill-defined on long aspect ratios) and the 2D block sampler.

\textbf{Masking strategy.} For each line, we sample $M = 4$ potentially overlapping
target masks, following the multi-block strategy of I-JEPA. Each target mask is a single contiguous
1D span whose width, expressed as a fraction of $T$, is drawn uniformly from
$[0.15, 0.25]$, and whose left edge is drawn uniformly over the valid
positions. We then form a single \emph{context mask} by taking the
complement of the union of the four target masks. Because target spans
overlap freely, the context can be small (we observed roughly $30$--$50\%$
of the line in expectation); we do not enforce a minimum context size.
This differs from canonical I-JEPA in two main ways: (i) target spans are sampled along a single spatial dimension
rather than as 2D rectangles, and (ii) the context is defined as the complement of the target
spans rather than sampled as a separate block. Bidirectional self-attention is used inside both the encoder and the
predictor; we did not find a benefit from causal masking on the context
side, consistent with the original I-JEPA results.

\textbf{Predictor.}
The predictor is a six-block transformer of hidden dimension $384$ and
six attention heads (about $11.6$M parameters at base). It receives the
context tokens projected from the encoder's hidden dimension into the
predictor's $384$-dimensional space, followed by one learnable
\texttt{[MASK]} token per masked target patch with the corresponding 1D
RoPE position. The full sequence (context tokens plus target placeholders)
is processed jointly so masked positions can attend to the available
context. The predictor's output at each target placeholder is then mapped
back to the encoder's hidden dimension with a single linear projection
before the loss.

\textbf{Targets.}
Target representations come from an exponential moving average copy of the
encoder, updated each step as $\bar\theta \leftarrow m\,\bar\theta + (1-m)\,\theta$
with a constant momentum $m = 0.9999$ (we do not ramp $m$). Targets are
extracted from the EMA encoder's \emph{final} layer, then
instance-normalised along the embedding dimension before the loss; this
prevents trivial solutions in which the predictor matches feature scale
rather than direction.

\textbf{Instance normalisation matters.} Per-patch instance
normalisation of the EMA targets is essential: without it, both I-JEPA and
V-JEPA-2 collapse early in training on HTR data. The loss saturates near
zero within a few thousand steps, the predictor learns the trivial
statistics of the target distribution (mean and scale of the embedding
vector), and the encoder's outputs become low-rank and effectively
constant across patches. We observed this collapse consistently on both
synthetic and real handwriting; instance-normalising targets along the
embedding dimension before computing the loss eliminates the collapse and
recovers stable training. We report all I-JEPA and V-JEPA-2 numbers with
instance normalisation enabled.

\textbf{Loss.}
The training objective is the L1 distance between predictor outputs and
EMA targets, summed over masked positions and averaged first within each
target mask and then across the four target masks of an image. We use L1
rather than smooth-L1 (the original I-JEPA's choice); we did not find a
difference at this scale. Gradients flow only through the online encoder
and predictor; the EMA target encoder receives no gradients (stop-gradient).

\textbf{Optimisation.} Same as the shared SSL recipe of
Sec.~\ref{app:ssl-pretraining}: AdamW, peak lr $3\times10^{-4}$, $10{,}000$
warmup steps, cosine decay, $500$ epochs of $2{,}000$ steps each.

\subsection{V-JEPA-2.}
\label{app:vjepa2-recipe}
V-JEPA-2 \citep{assran2025vjepa2} sits in the same JEPA framework as I-JEPA
and inherits everything from Sec.~\ref{app:ijepa-recipe}: the 1D
vertical-patch input geometry, the multi-block masking strategy with four
independent contiguous target spans of width drawn uniformly from
$[0.15, 0.25]$, the bidirectional encoder, the EMA target encoder with
constant momentum $m = 0.999$, instance-normalised targets, and the L1
loss aggregated within and across target masks. Despite its name we do not
treat the line image as a video; the input is still the 1D sequence of
vertical patches, since ``frames'' would not be meaningful for a single
static line image. Three deviations from I-JEPA distinguish V-JEPA-2.

\textbf{Multi-layer concatenated targets.}
Where I-JEPA matches the EMA encoder's final-layer feature at each masked
patch, V-JEPA-2 matches the channel-wise concatenation of the EMA encoder's
last four transformer-block outputs (layers $9$ through $12$ of the
$12$-block base encoder). Each layer's output is instance-normalised along
its embedding dimension separately before concatenation, so the four
sub-targets contribute on the same scale. The predictor's output dimension
is $4 \times d_{\text{enc}}$ rather than $d_{\text{enc}}$, and the L1 loss is
computed against the full concatenated target. This roughly quadruples the
target dimensionality and provides denser supervision per masked patch at
the cost of a larger predictor output projection.

\textbf{Deeper predictor.}
The predictor is a twelve-block transformer of hidden dimension $384$ and
six attention heads (about $17.0$M parameters at base), doubled in depth
from I-JEPA's six-block predictor. Predictor input format (context tokens
plus per-target learnable \texttt{[MASK]} tokens with 1D RoPE positions)
and back-projection to the target dimension are unchanged.

\textbf{Auxiliary context-side loss.}
A small auxiliary L1 loss is applied at the unmasked context positions: the
predictor sees the context and is also asked to reproduce the EMA targets
\emph{at those same context positions} (rather than at the masked target
positions only). The auxiliary loss is computed against the same
multi-layer concatenated EMA targets and weighted by a position-distance
factor that gives more weight to context positions adjacent to masked
spans. We linearly warm up the auxiliary-loss weight from zero over an
early training window so the main JEPA prediction objective stabilises
first; the warm-up window length matches the LR warmup window used by the
optimiser. Apart from these three changes, masking, EMA, and optimisation
are identical to I-JEPA (Sec.~\ref{app:ijepa-recipe}).

\subsection{SigLIP.}
\label{app:siglip-recipe}
SigLIP is a sigmoid contrastive image and text pretraining family
\citep{zhai2023siglip}. The image encoder is the same vertical-patch ViT as
the MIM and JEPA families, with patch features mean-pooled across the 256
patches into a single image embedding. The text encoder is a six-layer causal
transformer of the same width as the image encoder, operating on BPE-8192
tokenised text; the final token's hidden state is the text embedding. Both
modalities are linearly projected to a shared 256-dimensional space and
L2-normalised, and similarity is the dot product scaled by a learned
temperature. At base, the image encoder, text encoder, and projection heads
together total about $228$M parameters.

The training objective is the SigLIP loss \citep{zhai2023siglip}, namely a
per-pair sigmoid cross-entropy with positives on the diagonal of the batch
similarity matrix and a learned temperature scalar initialised at $0.07$ and
clamped to a maximum of $100$. Each line image is paired with its
ground-truth transcript; when augmentation is enabled the image is augmented
while the text is the unmodified transcript, and we do not mine hard
negatives or shuffle captions. The optimiser recipe matches the MIM family
with a higher peak learning rate of $6 \times 10^{-4}$.

\subsection{MoCo-v3 (sequential variant).}
\label{app:mocov3-recipe}
MoCo-v3 is a momentum contrastive image-only pretraining family
\citep{chen2021mocov3}. We adapt it to sequential line images via a
window-level positive pairing. Each image yields two augmented views (the
same augmentation pipeline applied twice independently), and each view is
divided into four contiguous windows along the time axis. The positive for a
given window in one view is the window at the same horizontal position in
the other view, while all other windows in the batch serve as negatives.
This is the digital-image-grounded variant of MoCo-v3 with a sequential
prior, and it allows contrastive learning on long line images without
collapsing to image-level matching.

The projection head is a three-layer MLP with hidden dimension $4096$ and
output dimension $256$, with batch normalisation and GELU on the hidden
layers, a linear last layer, and L2-normalised outputs. An EMA copy of the
encoder and projection head with momentum near $0.999$ serves as the
momentum encoder; negatives come from the current batch and we use no
separate queue. The temperature is $0.2$ for the real regime and $1.0$ for
the synth regime: the larger and more uniform synth distribution produces
more easily separable embeddings, and we observed early-training collapse at
$\tau = 0.2$ on synth (loss saturates near zero within a few thousand steps);
$\tau = 1.0$ was the smallest temperature in $\{0.2, 0.5, 1.0\}$ that avoided
this collapse, and we did not tune further. The training objective is the
InfoNCE loss between the projected query (online encoder) and the projected
key (momentum encoder) at matched windows, symmetrised across the two views.

\section{Handwritten Text Recognition datasets}
\label{app:datasets}

We evaluate on six benchmarks spanning five languages and several centuries
(Table~\ref{tab:datasets}). All datasets use the canonical train,
validation, and test splits from prior published work, and test data never
enters PCA fitting, encoder pretraining, probe training, or VLM fine-tuning.

\begin{table}[h]
\centering
\caption{HTR benchmarks used for evaluation. Splits are the canonical ones
from prior published work; line counts are reported in thousands.}
\label{tab:datasets}
\setlength{\tabcolsep}{6pt}
\renewcommand{\arraystretch}{1.15}
\footnotesize
\begin{tabular}{@{}lll rrr@{}}
\toprule
Dataset    & Language           & Period             & Train  & Val   & Test  \\
\midrule
IAM        & English            & modern             & 11.5K  & 1.1K  & 2.9K  \\
Rimes      & French             & modern             & 10.5K  & 1.0K  & 0.8K  \\
Bentham    & English            & 19th century       &  9.2K  & 1.4K  & 0.9K  \\
LAM        & Italian            & 17th--19th century & 19.8K  & 2.5K  & 2.5K  \\
Rodrigo    & Spanish            & 16th century       &  9.0K  & 1.0K  & 0.5K  \\
Parzival   & Middle High German & 13th century       &  2.2K  & 0.3K  & 1.3K  \\
\bottomrule
\end{tabular}
\end{table}

\paragraph{Synthetic data}
The synthetic pretraining corpus consists of roughly 12.5M handwriting
line images at the same resolution, rendered from Gutenberg \cite{gutenberg_dataset} text
in five languages (English, French, German, Spanish, Italian; about 2.5M lines per language). The generation pipeline varies font (about a
thousand handwriting fonts), font size, ink intensity, paper background,
slant, baseline jitter, and inter-word spacing. Synthetic data is used only
for pretraining encoders and alignment for encoder-decoder. 

\begin{figure}[h]
\centering
\includegraphics[width=\linewidth]{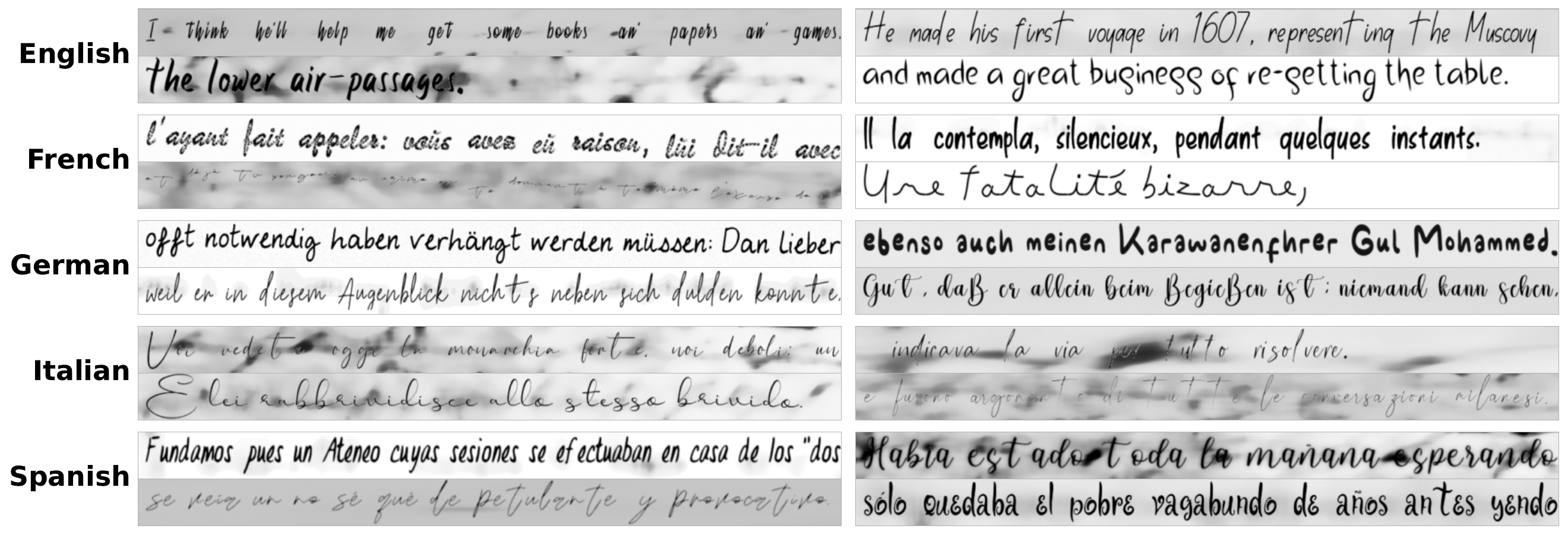}
\caption{\textbf{Synthetic handwriting samples used for SSL pretraining.} Two examples per language (English, French, German, Italian, Spanish) drawn from our synthetic corpus, rendered from Gutenberg text with handwriting fonts and per-line variation in font, ink intensity, paper background, slant, baseline jitter, and inter-word spacing. Each sample is a $64 \times 1024$ grayscale line image, the same resolution used by all SSL methods during pretraining.}
\label{fig:synth-samples}
\end{figure}

\section{CTC probes}
\label{app:ctc-probe}

We use two CTC probes throughout the paper, both applied on top of frozen
encoder features and operating at the \emph{character} level (vocabulary
size $|\mathcal{V}| = 94$ characters plus one CTC blank symbol; the same
character tokenizer is used for every dataset). Each probe emits one logit
vector per encoder patch, so the CTC alignment is per-frame and CER is
computed directly on the decoded string.

The Linear-CTC probe is a single linear projection from frozen encoder
embeddings to logits per patch, with no nonlinearity, no recurrence, and no
attention; it has zero sequence-modeling capacity by design and serves as
a readout-bottleneck test of how decodable each patch's feature is on its
own. With $D \cdot 95$ parameters, this is about 
$73$K at base
The BiLSTM-CTC probe adds a single
bidirectional LSTM layer with hidden size 256 between the frozen encoder
and the linear classifier; it has just enough sequential capacity to
bridge a non-text-aligned encoder feature sequence to a CTC-decodable
output, while still being shallow enough that its score reflects the
encoder rather than the head. The BiLSTM-CTC probe contains about 
$2.15$M at base size.

Both probes are trained with Adam at learning rate $10^{-3}$ (no weight
decay), gradient clipping at $1.0$, batch size 64, for 100 epochs without
early stopping, and we report the best validation CER reached during
training. The CTC blank index is the vocabulary size, decoding is greedy
CTC at the character level, and the encoder is frozen with EMA weights
loaded when the family provides them. We use the official train,
validation, and test splits, image size $1024 \times 64$, no augmentation
at probe time, and a single fixed random seed.

\paragraph{Reading the rescue gap.}
The Linear-CTC minus BiLSTM-CTC CER gap measures how much sequential
modelling capacity the BiLSTM adds on top of the encoder's features. A small
gap means the encoder features are already text-aligned (the linear head
suffices), and a large gap means the encoder features need sequence
modelling to be decodable.

\section{Evaluation with limited labels}
\label{app:low-label}

\begin{figure*}[h]
  \centering
  \includegraphics[width=1.0\textwidth]{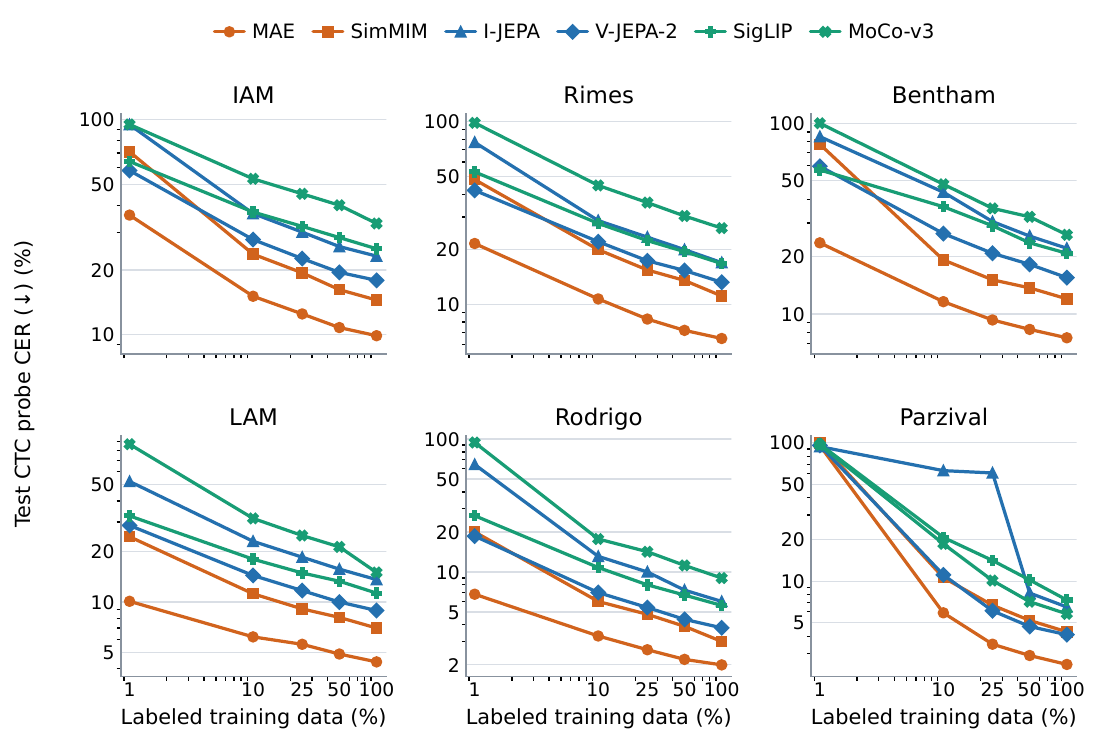}
    \caption{\textbf{Label efficiency on individual HTR benchmarks.}
    Test CER (\%) of BiLSTM CTC probes trained with 1\%, 10\%,
    25\%, 50\%, or 100\% of the labeled examples. Each curve uses a frozen
    encoder pretrained on real handwriting. Orange denotes pixel-ground methods,
    blue JEPA, and green contrastive objectives; markers identify individual
    methods.}
      \label{fig:low-label-all}
\end{figure*}

\paragraph{Protocol.} We freeze each encoder pretrained on real handwriting
and train the same BiLSTM readout with CTC loss on nested subsets containing
1\%, 10\%, 25\%, 50\%, or 100\% of each benchmark's labeled training
examples. All other probe settings follow the main experiments. We report
test character error rate (CER, \%) for every method, benchmark, and label
budget in Figure~\ref{fig:low-label-all}.

\paragraph{Results.} MAE has the lowest CER in 29 of the 30 benchmark and
label budget settings. The exception is Parzival at 1\% labels, where all
methods have high error and I-JEPA obtains 93.8\% CER compared with 95.6\%
for MAE. At budgets of 10\% or more, MAE and SimMIM have the two lowest CER
values on IAM, Rimes, Bentham, LAM, and Rodrigo. On Parzival, V-JEPA-2 ranks
second at 25\%, 50\%, and 100\% labels.

\section{VLM pipeline (Stages 1, 2, and 3)}
\label{app:vlm-pipeline}

For the HTR pipeline results we wrap the frozen SSL encoders into a
LLaVA-style decoder-only vision and language model. The frozen pretrained
SSL encoder (114M parameters at base) produces a sequence of
256 image patch features. A two-layer MLP modality projector (about 7.3M
parameters at base) maps encoder features to the decoder's hidden
dimension. The decoder is a pretrained causal language model with twelve
transformer layers, hidden dimension $1024$, sixteen heads, rotary
positions, and weight-tied token embedding (about 218M parameters at base);
it is pretrained on multilingual CC-100 text \cite{conneau-etal-2020-unsupervised-cc100} (English, Spanish, French,
German, Italian) under a standard next-token cross-entropy objective with
the same Byte-Pair Encoding with a vocab of 8192 used everywhere else. The BPE-8192 tokenizer is
built on the same multilingual CC-100 \cite{conneau-etal-2020-unsupervised-cc100} corpus before LM pretraining.

The decoder receives the projected image features as a left context (the
``image prefix'') followed by the BPE-tokenised transcript; attention is
causal, and the training loss is cross-entropy on transcript tokens only,
with positions in the image prefix masked from the loss. 

\paragraph{Stage 1 (alignment, synthetic only).}
Decoder warmstart with the encoder and the LM frozen and the projector
trained on synthetic data only, for 125 epochs of $2{,}000$ steps. This
stage acts as a label-supervised projector warm-start for Stages 2 and 3,
aligning the visual feature space with the LM embedding space.

\paragraph{Stage 2 (encoder-frozen fine-tuning).}
Encoder frozen, decoder unfrozen, projector unfrozen. Here we follow the probe-then-finetune \cite{liu2023llava,lin2024vila, liu2024nvila}. We per-dataset
fine-tune on the real training split for 40 epochs at a learning rate of
$5 \times 10^{-5}$ on the decoder and projection, with batch size 32. Per-dataset warmup steps reflect dataset size: roughly
$1{,}000$ for IAM, $1{,}700$ for Rimes, $1{,}400$ for Bentham, $600$ for
Parzival, $3{,}100$ for Rodrigo, and $3{,}900$ for LAM.

\paragraph{Stage 3 (full fine-tuning).}
Encoder unfrozen with a low encoder learning-rate multiplier of $0.05$,
decoder learning rate $2 \times 10^{-6}$, projection $5 \times 10^{-6}$,
peak base learning rate $5 \times 10^{-6}$. We train for 20 epochs with
no teacher-forcing noise, no label smoothing, and real augmentation enabled.
The optimiser is reinitialised from scratch (a fresh schedule starting from
the Stage 2 checkpoint).

\section{Evaluation protocol and baselines}
\label{app:evaluation}

We evaluate using Character Error Rate (CER), defined as the total Levenshtein
edit distance between predictions and references divided by the total number
of reference characters across the test set; lower is better. For both probes and
the VLM pipeline we pick the checkpoint that minimises validation CER and
report that checkpoint's test CER. Decoding is greedy CTC for the probes and
greedy autoregressive decoding (no beam search) for the VLM, and we did not
tune decoding hyperparameters. Probe and VLM evaluation are reported on the raw transcript with no case
folding and no punctuation stripping. 

\paragraph{Supervised baselines.} The supervised baselines in Table~\ref{tab:pipeline_sota} differ in their pretraining starting points, which we disclose here for transparency. \textbf{TrOCR-B}~\cite{trocr_li_2023} is initialised from the publicly released TrOCR-B checkpoint, which was pretrained on roughly 684M synthetic printed-text line images derived from PDF documents and Wikipedia text, and is then fine-tuned on our synthetic corpus and on each real dataset under our matched protocol. Our synthetic corpus contains $12.5$M handwriting line images, so TrOCR-B begins our pipeline with substantially more pretraining exposure than the SSL methods, which see only the unlabeled real handwriting at SSL time and our $12.5$M synthetic lines at the supervised stage. \textbf{DTrOCR}~\cite{dtrocr_fujitake_2023} does not have a publicly released checkpoint, so we initialise it from scratch and train it under the same matched protocol (full SFT on our synthetic corpus, then per-dataset full fine-tuning on real data). \textbf{CRNN}~\cite{multidimensional_recurrent_puig_2017} is also trained from scratch under the matched protocol. The SSL methods and the no-SSL Random init.\ row are likewise trained from scratch using the common backbone except for the SSL pretraining stage, which uses only the unlabeled real handwriting.

\section{Compute and reproducibility}
\label{app:compute}

\paragraph{Hardware.}
Experiments were carried out on one RTX 5090 ($32$ GB). Pretraining is
single-GPU throughout, with no data-parallel or model-parallel training, and
all training is in bfloat16 mixed precision.

\paragraph{Total compute.}
SSL pretraining across six methods and two data regimes required
approximately $1{,}500$ GPU-hours; CTC probes (12 encoders,
six datasets, two probes) required $300$ GPU-hours; the three-stage
VLM pipeline required $700$ GPU-hours; and PCA analyses and
variance-sweep probes required $30$ GPU-hours.
Total compute was approximately $2{,}500$ GPU-hours.

\stopcontents[appendix]

\end{document}